%% file: main.tex
\documentclass[10pt,twocolumn,letterpaper]{article}
\usepackage[T1]{fontenc}       %
\usepackage[pagenumbers]{cvpr}  %

\input{preamble}

\usepackage{amsmath,amssymb,bm}
\usepackage{booktabs,multirow,makecell}
\usepackage{graphicx}
\usepackage{algorithm}
\usepackage[noend]{algpseudocode}
\usepackage{placeins}
\usepackage{xspace}
\usepackage{amsmath} 
\newcommand{\RaTW}{RaTW\xspace}

\newcommand{\RFpp}{ReForge++\xspace}

\DeclareMathOperator*{\argmin}{arg\,min}
\definecolor{cvprblue}{rgb}{0.21,0.49,0.74}
\usepackage[pagebackref,breaklinks,colorlinks,allcolors=cvprblue]{hyperref}

\def\confName{CVPR}
\def\confYear{2027}

\title{Large-scale Model Merging via Recovery-aware Task Weighting}
\author{Kaiyang Li$^{1}$\quad Shaobo Han$^{2}$\quad Boming Chen$^{1}$\quad Martin Renqiang Min$^{2}$\quad Shihao Ji$^{1}$\\
$^{1}$University of Connecticut\quad $^{2}$NEC Laboratories America\\
{\tt\small \{kaiyang.li, boming.chen, shihao.ji\}@uconn.edu, \{shaobo, renqiang\}@nec-labs.com}}

\begin{document}
\maketitle
\raggedbottom
\input{sec/0_abstract}
\input{sec/1_intro}
\input{sec/fig2}
\input{sec/2_related}
\input{sec/3_method}

\input{sec/4_experiments}

\input{sec/5_conclusion}

\section*{LLM Usage Statement}
We used OpenAI Codex and Anthropic Claude to assist with code implementation and review, and to polish the language of the manuscript.

{
    \small
    \bibliographystyle{ieeenat_fullname}
    \bibliography{main}
}
\clearpage
\setcounter{equation}{0}
\maketitlesupplementary
\appendix
\renewcommand{\theequation}{A\arabic{equation}}\renewcommand{\theHequation}{A\arabic{equation}}
\setcounter{figure}{0}\setcounter{table}{0}
\renewcommand{\thefigure}{A\arabic{figure}}\renewcommand{\theHfigure}{A\arabic{figure}}
\renewcommand{\thetable}{A\arabic{table}}\renewcommand{\theHtable}{A\arabic{table}}
\input{sec/X_appendix}

\end{document}

%% file: preamble.tex
\newcommand{\DFs}{DF-stat\xspace}
\newcommand{\DAs}{DA-stat\xspace}

\usepackage{microtype}
\LoadMicrotypeFile{ptm}
\SetProtrusion[name=ptm-T1-noplus,load=ptm-T1]{encoding=T1,family=ptm}{+ = {0,0}}

%% file: sec/0_abstract.tex
\begin{abstract}
Model merging combines multiple task-specific expert models into a single model. However, merging preserves expert capabilities unevenly across tasks. This imbalance is particularly pronounced in large-scale settings, where some tasks can suffer severe capability losses even when average accuracy remains high. To address this imbalance, we introduce Recovery-aware Task Weighting (\RaTW), a model merging framework that assigns larger weights to tasks whose expert capabilities are poorly retained by an initial merged model. We quantify these recovery opportunities as the fraction of a task's labeled validation samples that its expert predicts correctly but the initial merged model gets wrong, and use the resulting task weights to re-merge the experts. To evaluate merging at large scale, we construct CLIPMerge-100, a 100-task benchmark with task-specific experts on CLIP ViT-L/14 and ViT-B/32. Across eleven merging methods with 20, 50, and 100 tasks on two CLIP backbones, \RaTW improves both mean accuracy and tail20 (the mean expert-normalized accuracy of the 20\% worst-retained tasks) in all 33 ViT-L/14 settings and 32 of 33 ViT-B/32 settings. We further introduce \RFpp, which combines \RaTW with iterative model refinement and outperforms all evaluated baselines, including a matched-budget control, in both mean accuracy and tail20 across all settings.
\end{abstract}

%% file: sec/1_intro.tex
\section{Introduction}
\label{sec:intro}

Model merging combines task-specific expert models into a single model without joint training~\cite{ilharco2023editing,jin2023regmean}. However, merging preserves expert capabilities unevenly across tasks, particularly in large-scale settings~\cite{marczak2025isoc}, where some tasks suffer substantial capability losses even when average accuracy remains high. To evaluate retention on these poorly preserved tasks, we define \emph{tail20} as the average ratio of merged-model accuracies to expert-model accuracies over the worst-retained 20\% of tasks. Figure~\ref{fig:teaser}(a) shows that across the eleven evaluated methods tail20 is only 12\%--39\% when merging 100 tasks on ViT-L/14.

\input{sec/fig1}

This raises a central question: can we recover capabilities on poorly retained tasks, while maintaining or even improving overall performance? We study this question in the setting of \emph{model merging with validation feedback}. Merging methods commonly use a labeled validation set to select hyperparameters; we additionally use it to measure what the current merge fails to retain. We address this question through task weighting, which controls how much each expert contributes to the merged model. Most existing model-merging methods weight tasks equally~\cite{ilharco2023editing} or derive their contributions from parameter or feature statistics~\cite{matena2022merging,jin2023regmean}, without directly accounting for what the current merge has failed to retain. Those that do use such feedback either learn coefficients by backpropagation~\cite{yang2024adamerging,zhang2026expertmerging} or interleave merging with expert retraining~\cite{metis2026}. We introduce Recovery-aware Task Weighting (\RaTW), a plug-and-play method that adjusts task weights based on how much of each expert's capability is lost after merging. Specifically, upon obtaining an initial merged model, we quantify this loss as the fraction of each task's labeled validation samples that its expert predicts correctly but the merged model gets wrong. \RaTW then merges the experts again, assigning higher weights to tasks with larger capability losses (Fig.~\ref{fig:overview}(a)). Most model-merging methods fall into two broad categories: aggregation-based and reconstruction-based; \RaTW is designed to work with both.

We further introduce \RFpp, which integrates \RaTW with ReForge~\cite{li2026bmm}. ReForge treats an existing merged model as a prior and refines it module by module with evidence observed from the experts, namely their activations or task vectors, to obtain a posterior merged model. \RFpp makes this refinement recovery-aware: each task's evidence is weighted by its \RaTW weight, so that poorly retained tasks receive more emphasis. It is also iterative: each posterior becomes the prior for the next round, where the task weights are recomputed (Fig.~\ref{fig:overview}(b)), so that each round typically starts from a better merged model.

To support large-scale model-merging evaluation, we also construct CLIPMerge-100, a benchmark spanning 100 tasks, and train a task-specific expert for each task on both CLIP ViT-L/14 and ViT-B/32. Experiments on this benchmark evaluate \RaTW across eleven merging methods at 20, 50, and 100 tasks, using both mean accuracy and tail20. \RaTW improves both metrics in all 33 ViT-L/14 settings (Fig.~\ref{fig:teaser}(b)) and 32 of 33 ViT-B/32 settings. On two vision-language models, \RaTW also improves both metrics for all five evaluated merging methods (Sec.~\ref{sec:vlm}). Furthermore, \RFpp, our iterative extension of ReForge, outperforms all evaluated baselines, including a matched-budget control, in both mean accuracy and tail20 across all settings (Sec.~\ref{sec:main}). Controlled experiments under fixed merging configurations and search budgets further examine the contribution of recovery-aware task weighting. Our contributions are threefold:
\begin{itemize}\setlength{\itemsep}{0pt}\setlength{\parskip}{0pt}
  \item Recovery-aware Task Weighting (\RaTW), a plug-and-play, backpropagation-free method that assigns larger weights to tasks whose expert capabilities are poorly retained by the merged model.
  \item \RFpp, which makes ReForge's prior-to-posterior refinement recovery-aware and iterative.
  \item CLIPMerge-100, a large-scale model-merging benchmark on 100 tasks with task-specific experts trained on both CLIP ViT-L/14 and ViT-B/32, together with extensive evaluations of average performance and tail-task retention, controlled experiments, and task-level analyses.
\end{itemize}

%% file: sec/fig1.tex
\begin{figure*}[t]
  \centering
  \includegraphics[width=\textwidth]{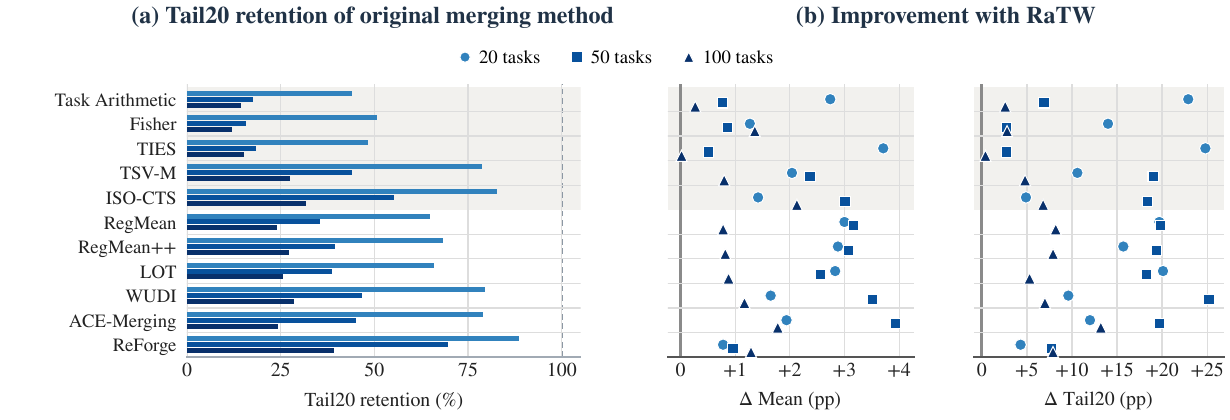}
  \caption{\textbf{Expert capability retention and gains from \RaTW (ViT-L/14).}
  Both panels cover eleven merging methods at $T=20$, 50 and 100 tasks (33 settings).
  (a) Tail20 retention of the original merging methods, i.e., the mean expert-normalized accuracy of the 20\% worst-retained tasks; 100\% denotes expert-level performance.
  (b) Gains in mean accuracy and tail20 (percentage points) after integrating \RaTW into each merging method.
  ReForge is run for one round with data-assisted statistics (\DAs; Sec.~\ref{sec:rw}), starting from the ISO-CTS merge. Gray-shaded rows indicate aggregation-based methods, while unshaded rows indicate reconstruction-based methods.}
  \label{fig:teaser}
\end{figure*}

%% file: sec/fig2.tex
\begin{figure*}[t]
  \centering
  \includegraphics[width=\textwidth]{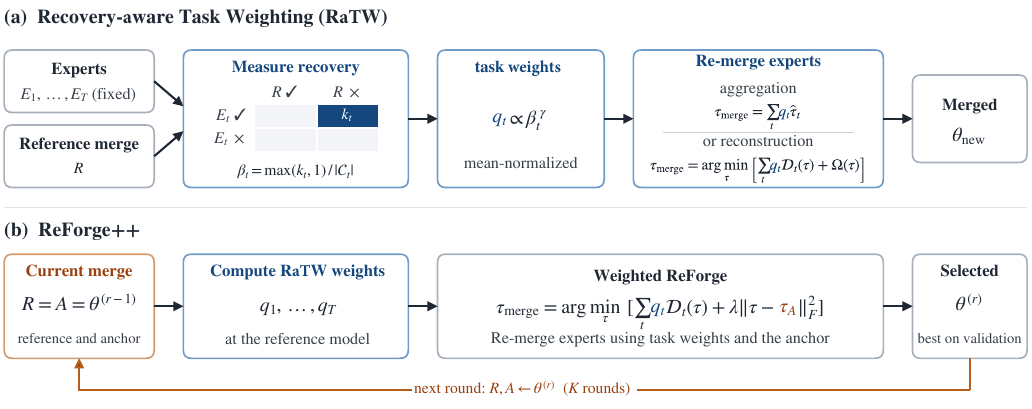}
  \caption{\textbf{Overview of \RaTW and \RFpp.} (a) \RaTW measures capability losses using validation samples that experts predict correctly but the reference model gets wrong. Tasks with larger losses receive higher weights in either aggregation-based or reconstruction-based merging methods. Here, $\theta_0$ is the pre-trained model and $\widehat{\tau}_t$ denotes a task vector after method-specific processing and normalization. Both paradigms produce a merged update $\tau_{\mathrm{merge}}$, which is added to $\theta_0$ to yield a new merged model $\theta_{\rm{new}}$. (b) In each \RFpp round, \RaTW computes task weights from the reference model. ReForge merges the experts using these weights and the reference model as its anchor. The resulting model becomes both the reference model and the anchor for the next round. \textcolor[rgb]{0.075,0.28,0.49}{Blue} marks the components specific to \RaTW, and \textcolor[rgb]{0.6,0.25,0.055}{orange} marks those specific to \RFpp.}
  \label{fig:overview}
\end{figure*}

%% file: sec/2_related.tex
\section{Related Work}
\label{sec:related}

\subsection{Model Merging}
\label{sec:rel-agg}

Most model-merging methods can be broadly grouped into two categories: aggregation-based and reconstruction-based methods. Aggregation-based methods construct a merged update by summing original or processed task vectors. Task Arithmetic~\cite{ilharco2023editing} adds scaled task vectors to the pre-trained model, while subsequent methods trim~\cite{yadav2023ties}, mask~\cite{wang2024consensus,davari2024breadcrumbs}, or decorrelate~\cite{gargiulo2025tsv} task updates to reduce interference. ISO-CTS~\cite{marczak2025isoc} flattens the singular-value spectrum of the merged update and incorporates task-specific subspaces. Fisher merging~\cite{matena2022merging} averages expert parameters using element-wise weights derived from diagonal Fisher information.

Reconstruction-based methods estimate merged parameters by fitting expert-specific parameters or responses, typically through regression. RegMean~\cite{jin2023regmean} matches expert linear-layer outputs. RegMean++~\cite{nguyen2025regmeanpp} and Chain of Merges~\cite{chainofmerges2025} account for activations propagated through already merged layers, and LOT-Merging~\cite{sun2025lot} minimizes layer-wise feature drift. WUDI-Merging~\cite{cheng2025wudi} uses task vectors as input proxies and normalizes task terms by their squared norms; ACE-Merging~\cite{ace2025} constructs a merging system from task-vector covariances. ReForge~\cite{li2026bmm} treats an existing merged model as a prior and refines it with evidence observed from the experts, namely their activations or task vectors, to obtain a posterior merged model. As a plug-and-play method, \RaTW can be integrated into both aggregation-based and reconstruction-based merging by reweighting task vectors and task-specific fitting losses, respectively. In particular, integrating \RaTW with ReForge~\cite{li2026bmm} yields \RFpp, which reweights this evidence by recovery scores and iterates the refinement, using each posterior as the prior for the next round.

\subsection{Task Weighting}
\label{sec:rel-prio}

Several methods have been proposed to determine model-merging weights. AdaMerging~\cite{yang2024adamerging} and Expert Merging~\cite{zhang2026expertmerging} learn merging coefficients from unlabeled data through gradient-based optimization. AdaMerging minimizes prediction entropy, while Expert Merging aligns the merged model's representations and predictions with those of the experts. Coefficients can also be estimated directly from model parameters: MetaGPT~\cite{zhou2024metagpt} uses squared task-vector norms, while NAN~\cite{si2025nan} uses inverse parameter norms. MAP~\cite{li2025map} evaluates candidate merges and uses a quadratic approximation of task performance to search for merging coefficients. Unlike methods that merge fixed experts, METIS~\cite{metis2026} alternates local training and merging. It assigns task weights according to the loss gaps between the current merged model and the locally adapted experts. In contrast, \RaTW determines task weights using a recovery signal. For each task, it measures the fraction of its validation samples that its expert predicts correctly but the merged model gets wrong. These weights are computed directly from sample-level counts, requiring neither gradient-based coefficient learning nor expert retraining. Upweighting hard examples or tasks is a long-standing idea, from boosting~\cite{freund1997decision} to GradNorm-style multi-task loss balancing~\cite{chen2018gradnorm}, dynamic task prioritization~\cite{guo2018dtp} and group distributionally robust optimization~\cite{sagawa2020dro}. These methods upweight what is hard for the model being trained; \RaTW instead upweights tasks whose capabilities the experts already have but the merged model has lost, and applies these weights when merging fixed experts.

%% file: sec/3_method.tex
\section{Our Method}~\label{sec:method}
This section defines the recovery score of each task (Sec.~\ref{sec:est}), applies the resulting task weights to aggregation-based and reconstruction-based merging (Sec.~\ref{sec:rw}), and combines \RaTW with ReForge into \RFpp (Sec.~\ref{sec:rfpp}).

\subsection{Estimating Recovery Opportunities}~\label{sec:est}
We consider merging $T$ task-specific experts $\{E_t\}_{t=1}^{T}$, each fine-tuned from a shared pre-trained model $\theta_0$. We first apply an existing merging method (e.g., Task Arithmetic~\cite{ilharco2023editing}) to obtain a reference model $R$. For each task $t$, we evaluate $R$ and the expert $E_t$ on the same labeled validation set $\mathcal C_t$. Samples that $E_t$ predicts correctly but $R$ gets wrong reveal expert capabilities lost during merging. With $k_t$ denoting the number of such samples,
\begin{equation}
  k_t=\sum_{(x,y)\in\mathcal C_t}\mathbf 1\!\left[E_t(x)=y,\;R(x)\neq y\right],\nonumber
\end{equation}
the recovery score is defined as
\begin{equation}
  \beta_t=\frac{\max(k_t,1)}{|\mathcal C_t|}.~\label{eq:beta}
\end{equation}
The ratio $k_t/|\mathcal C_t|$ is the fraction of validation samples on which the expert is correct but $R$ is wrong. \RaTW uses this signal to prioritize recovery opportunities across tasks. When $k_t=0$, the minimum count of one prevents the task from receiving zero weight and being excluded from re-merging, since observing no recovery opportunities on the validation set does not imply that the task can be ignored.

\subsection{Recovery-Weighted Merging}
\label{sec:rw}
We can then convert recovery scores into task weights by:
\begin{equation}
  q_t=\frac{\beta_t^{\gamma}}{\frac{1}{T}\sum_{j=1}^{T}\beta_j^{\gamma}}, \;\;\;\;\textrm{with } \gamma\geq0.~\label{eq:q}
\end{equation}
Let $\mathbf q=(q_1,\ldots,q_T)^\top$ denote the vector of task weights. Larger $\gamma$ places greater emphasis on tasks with more capability loss, while $\gamma=0$ gives uniform weights. The exponent $\gamma$ is selected on the validation set together with each method's own hyperparameters (Sec.~\ref{sec:setup}). The same $T$ experts are then re-merged with weights $\mathbf q$ by either weighted aggregation or weighted reconstruction (Fig.~\ref{fig:overview}(a)).

\paragraph{Weighted aggregation.}
Task Arithmetic~\cite{ilharco2023editing}, Fisher merging~\cite{matena2022merging}, TIES~\cite{yadav2023ties}, TSV-M~\cite{gargiulo2025tsv}, and ISO-CTS~\cite{marczak2025isoc} are the aggregation-based merging methods. Let $\theta_t$ denote expert $t$'s parameters and $\tau_t=\theta_t-\theta_0$ its task vector. \RaTW forms a weighted sum of their original or processed task vectors by:
\[
  \tau_{\mathrm{merge}}(\mathbf q)=\sum_{t=1}^{T}q_t\widehat{\tau}_t(\mathbf q).
\]
Here, $\widehat{\tau}_t(\mathbf q)$ denotes the task vector after method-specific processing and normalization. The processed contribution may depend on $\mathbf q$ through weighted normalization, as in Fisher merging and TIES. For Task Arithmetic~\cite{ilharco2023editing}, $\widehat{\tau}_t(\mathbf q)=\tau_t$.

\begin{algorithm}[t]
  \caption{\RFpp}
  \label{alg:rfpp}
  \small
  \begin{algorithmic}[1]
    \Require experts and their statistics $\{G_t\}$; initial model $\theta^{(0)}$; labeled validation data; $K$ rounds; $B$ Bayesian optimization (BO) evaluations per round
    \For{$r=1,\dots,K$}
      \State $R,A\gets\theta^{(r-1)}$
      \State Compute and fix $\beta_t$ for this round (Eq.~\eqref{eq:beta})
      \State Initialize BO for this round
      \For{$j=1,\dots,B$}
        \State BO proposes $(\mathbf s_j,\boldsymbol\lambda_j,\gamma_j)$
        \State Compute $\mathbf q_j$ from $\{\beta_t\}$ and $\gamma_j$ (Eq.~\eqref{eq:q})
        \State Build $\widetilde\theta_j$ with recovery-weighted ReForge using anchor $A$, weights $\mathbf q_j$, regularization $\boldsymbol\lambda_j$, and scales $\mathbf s_j$
        \State $v_j\gets$ mean validation accuracy of $\widetilde\theta_j$; report $v_j$ to BO
      \EndFor
      \State $j^\star\gets\arg\max_j v_j$; $\theta^{(r)}\gets\widetilde\theta_{j^\star}$
    \EndFor
    \State \Return $\theta^{(K)}$
  \end{algorithmic}
\end{algorithm}
\paragraph{Weighted reconstruction.}
RegMean~\cite{jin2023regmean}, RegMean++~\cite{nguyen2025regmeanpp}, LOT-Merging~\cite{sun2025lot}, WUDI~\cite{cheng2025wudi}, ACE-Merging~\cite{ace2025}, and ReForge~\cite{li2026bmm} are the reconstruction-based merging methods. For these methods, \RaTW weights each task's fitting error:
\begin{equation}
  \tau_{\mathrm{merge}}=\argmin_{\tau}\left[\sum_{t=1}^{T}q_t\,\mathcal D_t(\tau)+\Omega(\tau)\right],
  \label{eq:obj}
\end{equation}
where $\mathcal D_t(\tau)$ measures the mismatch between the merged module and the corresponding module of expert $t$, such as their output difference on task $t$'s input, while $\Omega(\tau)$ is the method-specific regularization penalty. Eq.~\eqref{eq:obj} is solved separately for each module, and the same $q_t$ weights task $t$'s fitting term in every module.

For RegMean~\cite{jin2023regmean}, the regression fits each merged layer's outputs to the expert's outputs on its task, with no additional regularization term, $\Omega(\tau)\equiv0$. Its fitting error in task-vector coordinates is
\begin{equation}
  \mathcal D_t(\tau)=\mathrm{tr}\!\left[(\tau-\tau_t)G_t(\tau-\tau_t)^\top\right].
  \label{eq:quadratic_fit}
\end{equation}
We define $X_t$ by row-wise stacking the input activation vectors to expert $t$'s corresponding layer, and denote its number of rows by $N_t^{\mathrm{in}}$. We then compute
\[
  G_t=\frac{\alpha X_t^\top X_t+(1-\alpha)\operatorname{Diag}(X_t^\top X_t)}{N_t^{\mathrm{in}}},
\]
where $\operatorname{Diag}(\cdot)$ retains only the diagonal entries. The factor $\alpha$ regularizes the reconstruction by shrinking the off-diagonal entries of the Gram matrix.

For ReForge~\cite{li2026bmm}, an existing merged model $A$, called the anchor, serves as a prior and is refined by reconstructing its modules from the experts. ReForge uses the quadratic fitting form in Eq.~\eqref{eq:quadratic_fit}, with task-vector statistics for its data-free variant (\DFs) and activation statistics for its data-assisted variant (\DAs):
\[
  G_t^{\mathrm{DF}}=\frac{\tau_t^\top\tau_t}{\|\tau_t\|_F^2},\qquad
  G_t^{\mathrm{DA}}=\frac{X_t^\top X_t}{\operatorname{tr}(X_t^\top X_t)}.
\]
Here, $\tau_t$ is the expert's task-vector block for the module being reconstructed, and each row of $X_t$ is an input activation to that module on task $t$'s training data. To keep the reconstruction close to the anchor, ReForge adds the penalty
\[
  \Omega(\tau)=\lambda\|\tau-\tau_A\|_F^2,
\]
where $\tau_A=A-\theta_0$ is the anchor's task vector and $\lambda>0$ controls how strongly deviations from the anchor are penalized. The anchor can be any merged model produced by a merging method. After reconstruction, scaling factors control the magnitude of the merged updates. We collect these scaling factors in $\mathbf s$ and the regularization strengths in $\boldsymbol\lambda$. ReForge selects these hyperparameters by Bayesian optimization (BO)~\cite{snoek2012practical}, using mean validation accuracy to evaluate each candidate model.

Original ReForge uses uniform task weights, $q_t=1$. To apply \RaTW, we use the anchor as the reference, $R=A$, to measure which expert capabilities it has lost, then weight the fitting terms by $q_t$ in Eq.~\eqref{eq:obj}. Both \DFs and \DAs use labeled validation data to compute the recovery scores. The weighted forms and solvers of all eleven methods are given in App.~\ref{app:derive}.

\begin{table*}[!t]
  \centering
  \caption{\textbf{\RaTW across eleven merging methods on ViT-L/14.} Mean accuracy and tail20 (\%) are reported for merging 20, 50 and 100 task experts. Native denotes the original method; +\RaTW applies recovery-aware task weighting. $\Delta$ reports the gain over Native in percentage points. Bold marks the better result in each pair. Average is taken across the eleven methods. ReForge is run for one round with \DAs, starting from the ISO-CTS merge. Zero-shot CLIP and the average of the individual experts are listed for reference. }
  \label{tab:transfer}
  \footnotesize
  \renewcommand{\arraystretch}{1.12}
  \setlength{\aboverulesep}{0.3ex}\setlength{\belowrulesep}{0.5ex}
  \setlength{\tabcolsep}{1pt}
  \resizebox{\textwidth}{!}{%
  \begin{tabular}{@{}cl*{18}{r}@{}}
    \toprule
    & & \multicolumn{6}{c}{$T=20$} & \multicolumn{6}{c}{$T=50$} & \multicolumn{6}{c}{$T=100$} \\
    \cmidrule(lr){3-8}\cmidrule(lr){9-14}\cmidrule(l){15-20}
    & & \multicolumn{3}{c}{Mean} & \multicolumn{3}{c}{tail20} & \multicolumn{3}{c}{Mean} & \multicolumn{3}{c}{tail20} & \multicolumn{3}{c}{Mean} & \multicolumn{3}{c}{tail20} \\
    \cmidrule(lr){3-5}\cmidrule(lr){6-8}\cmidrule(lr){9-11}\cmidrule(lr){12-14}\cmidrule(lr){15-17}\cmidrule(l){18-20}
    & Method & \multicolumn{1}{c}{Native} & \multicolumn{1}{c}{+\RaTW} & \multicolumn{1}{c}{$\Delta$} & \multicolumn{1}{c}{Native} & \multicolumn{1}{c}{+\RaTW} & \multicolumn{1}{c}{$\Delta$} & \multicolumn{1}{c}{Native} & \multicolumn{1}{c}{+\RaTW} & \multicolumn{1}{c}{$\Delta$} & \multicolumn{1}{c}{Native} & \multicolumn{1}{c}{+\RaTW} & \multicolumn{1}{c}{$\Delta$} & \multicolumn{1}{c}{Native} & \multicolumn{1}{c}{+\RaTW} & \multicolumn{1}{c}{$\Delta$} & \multicolumn{1}{c}{Native} & \multicolumn{1}{c}{+\RaTW} & \multicolumn{1}{c}{$\Delta$} \\
    \midrule
     & \textit{Zero-shot CLIP} & 65.61 &  &  & 32.6 &  &  & 53.20 &  &  & 13.4 &  &  & 48.11 &  &  & 12.1 &  &  \\
     & \textit{Expert average} & 93.70 &  &  & 100.0 &  &  & 92.18 &  &  & 100.0 &  &  & 89.79 &  &  & 100.0 &  &  \\
    \midrule
    \multirow{5}{*}{\rotatebox[origin=c]{90}{\scriptsize Aggregation}} & Task Arithmetic~\cite{ilharco2023editing} & 73.30 & \textbf{76.04} & $+2.74$ & 44.0 & \textbf{66.9} & $+22.9$ & 58.82 & \textbf{59.59} & $+0.77$ & 17.5 & \textbf{24.3} & $+6.9$ & 51.64 & \textbf{51.91} & $+0.27$ & 14.3 & \textbf{16.9} & $+2.6$ \\
     & Fisher~\cite{matena2022merging} & 73.47 & \textbf{74.74} & $+1.27$ & 50.7 & \textbf{64.6} & $+14.0$ & 57.19 & \textbf{58.05} & $+0.86$ & 15.7 & \textbf{18.4} & $+2.7$ & 50.34 & \textbf{51.69} & $+1.35$ & 12.1 & \textbf{14.7} & $+2.7$ \\
     & TIES~\cite{yadav2023ties} & 74.92 & \textbf{78.63} & $+3.71$ & 48.4 & \textbf{73.2} & $+24.8$ & 59.20 & \textbf{59.73} & $+0.52$ & 18.4 & \textbf{21.1} & $+2.7$ & 52.21 & \textbf{52.24} & $+0.02$ & 15.2 & \textbf{15.7} & $+0.4$ \\
     & TSV-M~\cite{gargiulo2025tsv} & 86.26 & \textbf{88.30} & $+2.04$ & 78.6 & \textbf{89.2} & $+10.6$ & 73.74 & \textbf{76.11} & $+2.37$ & 44.1 & \textbf{63.1} & $+19.0$ & 64.74 & \textbf{65.54} & $+0.80$ & 27.4 & \textbf{32.2} & $+4.8$ \\
     & ISO-CTS~\cite{marczak2025isoc} & 87.75 & \textbf{89.17} & $+1.42$ & 82.7 & \textbf{87.6} & $+4.9$ & 78.27 & \textbf{81.28} & $+3.01$ & 55.3 & \textbf{73.8} & $+18.4$ & 68.14 & \textbf{70.27} & $+2.13$ & 31.8 & \textbf{38.6} & $+6.8$ \\
    \midrule
    \multirow{6}{*}{\rotatebox[origin=c]{90}{\scriptsize Reconstruction}} & RegMean~\cite{jin2023regmean} & 82.81 & \textbf{85.81} & $+3.00$ & 64.9 & \textbf{84.6} & $+19.7$ & 69.91 & \textbf{73.07} & $+3.16$ & 35.6 & \textbf{55.3} & $+19.8$ & 60.99 & \textbf{61.77} & $+0.78$ & 23.9 & \textbf{32.1} & $+8.2$ \\
     & RegMean++~\cite{nguyen2025regmeanpp} & 83.65 & \textbf{86.54} & $+2.88$ & 68.3 & \textbf{84.0} & $+15.7$ & 71.50 & \textbf{74.57} & $+3.07$ & 39.4 & \textbf{58.8} & $+19.4$ & 63.18 & \textbf{64.00} & $+0.82$ & 27.3 & \textbf{35.2} & $+7.9$ \\
     & LOT~\cite{sun2025lot} & 82.86 & \textbf{85.70} & $+2.83$ & 65.8 & \textbf{86.0} & $+20.1$ & 71.13 & \textbf{73.69} & $+2.56$ & 38.8 & \textbf{57.1} & $+18.3$ & 62.08 & \textbf{62.96} & $+0.88$ & 25.7 & \textbf{31.0} & $+5.3$ \\
     & WUDI~\cite{cheng2025wudi} & 87.62 & \textbf{89.27} & $+1.65$ & 79.6 & \textbf{89.1} & $+9.6$ & 76.03 & \textbf{79.54} & $+3.51$ & 46.7 & \textbf{71.9} & $+25.2$ & 66.35 & \textbf{67.52} & $+1.17$ & 28.5 & \textbf{35.5} & $+7.0$ \\
     & ACE-Merging~\cite{ace2025} & 87.36 & \textbf{89.30} & $+1.94$ & 79.1 & \textbf{91.1} & $+12.0$ & 73.57 & \textbf{77.52} & $+3.94$ & 45.0 & \textbf{64.7} & $+19.7$ & 62.05 & \textbf{63.84} & $+1.78$ & 24.3 & \textbf{37.5} & $+13.2$ \\
     & ReForge~\cite{li2026bmm} & 90.02 & \textbf{90.81} & $+0.78$ & 88.6 & \textbf{92.9} & $+4.3$ & 83.69 & \textbf{84.65} & $+0.96$ & 69.7 & \textbf{77.4} & $+7.7$ & 73.36 & \textbf{74.65} & $+1.29$ & 39.3 & \textbf{47.2} & $+7.9$ \\
    \midrule
    \multicolumn{2}{@{}l}{\textit{Average}} & 82.73 & \textbf{84.94} & $+2.21$ & 68.2 & \textbf{82.7} & $+14.4$ & 70.28 & \textbf{72.53} & $+2.25$ & 38.7 & \textbf{53.3} & $+14.5$ & 61.37 & \textbf{62.40} & $+1.03$ & 24.5 & \textbf{30.6} & $+6.1$ \\
    \bottomrule
  \end{tabular}%
  }
\end{table*}

\subsection{ReForge++: Iterative Refinement}
\label{sec:rfpp}

\RFpp performs $K$ rounds of recovery-weighted ReForge refinement, producing a sequence of merged models $\theta^{(1)},\ldots,\theta^{(K)}$ from an initial merged model $\theta^{(0)}$ (Fig.~\ref{fig:overview}(b); Alg.~\ref{alg:rfpp}). We iterate because the refined model typically improves on its prior, so it serves as a stronger prior for the next round. At the start of round $r$, we use the previous model as both the reference and the anchor, $R=A=\theta^{(r-1)}$, and compute the recovery scores $\beta_t$ once using Eq.~\eqref{eq:beta}; the anchor and scores stay fixed throughout the round. BO then runs $B$ evaluations. In the $j$-th evaluation, BO proposes the scaling factors $\mathbf s_j$, regularization strengths $\boldsymbol\lambda_j$, and recovery exponent $\gamma_j$. Because the recovery scores are fixed within the round, the task weights $\mathbf q_j$ vary across evaluations only through $\gamma_j$ (Eq.~\eqref{eq:q}). ReForge then builds the candidate model $\widetilde\theta_j$ with $\mathbf q_j$, $\boldsymbol\lambda_j$, and $\mathbf s_j$, and its mean validation accuracy is fed back to BO. After the $B$ evaluations, the candidate with the highest mean validation accuracy becomes $\theta^{(r)}$, which serves as the reference and anchor of the next round. After $K$ rounds, \RFpp returns $\theta^{(K)}$. We use $K=3$ by default (rounds are analyzed in Sec.~\ref{sec:budget}); the choice of $\theta^{(0)}$ and $B$ is described in Sec.~\ref{sec:exp}.

%% file: sec/4_experiments.tex
\input{sec/4a_exp}

\input{sec/4b_analysis}

%% file: sec/4a_exp.tex
\section{Experiments}
\label{sec:exp}

\begin{table*}[!t]
  \centering
  \caption{\textbf{\RaTW across eleven merging methods on ViT-B/32.} Mean accuracy and tail20 (\%) are reported for merging 20, 50 and 100 task experts. Notation follows Table~\ref{tab:transfer}.}
  \label{tab:transfer-b32}
  \footnotesize
  \renewcommand{\arraystretch}{1.12}
  \setlength{\aboverulesep}{0.3ex}\setlength{\belowrulesep}{0.5ex}
  \setlength{\tabcolsep}{1pt}
  \resizebox{\textwidth}{!}{%
  \begin{tabular}{@{}cl*{18}{r}@{}}
    \toprule
    & & \multicolumn{6}{c}{$T=20$} & \multicolumn{6}{c}{$T=50$} & \multicolumn{6}{c}{$T=100$} \\
    \cmidrule(lr){3-8}\cmidrule(lr){9-14}\cmidrule(l){15-20}
    & & \multicolumn{3}{c}{Mean} & \multicolumn{3}{c}{tail20} & \multicolumn{3}{c}{Mean} & \multicolumn{3}{c}{tail20} & \multicolumn{3}{c}{Mean} & \multicolumn{3}{c}{tail20} \\
    \cmidrule(lr){3-5}\cmidrule(lr){6-8}\cmidrule(lr){9-11}\cmidrule(lr){12-14}\cmidrule(lr){15-17}\cmidrule(l){18-20}
    & Method & \multicolumn{1}{c}{Native} & \multicolumn{1}{c}{+\RaTW} & \multicolumn{1}{c}{$\Delta$} & \multicolumn{1}{c}{Native} & \multicolumn{1}{c}{+\RaTW} & \multicolumn{1}{c}{$\Delta$} & \multicolumn{1}{c}{Native} & \multicolumn{1}{c}{+\RaTW} & \multicolumn{1}{c}{$\Delta$} & \multicolumn{1}{c}{Native} & \multicolumn{1}{c}{+\RaTW} & \multicolumn{1}{c}{$\Delta$} & \multicolumn{1}{c}{Native} & \multicolumn{1}{c}{+\RaTW} & \multicolumn{1}{c}{$\Delta$} & \multicolumn{1}{c}{Native} & \multicolumn{1}{c}{+\RaTW} & \multicolumn{1}{c}{$\Delta$} \\
    \midrule
     & \textit{Zero-shot CLIP} & 55.65 &  &  & 22.0 &  &  & 46.66 &  &  & 9.1 &  &  & 42.05 &  &  & 9.3 &  &  \\
     & \textit{Expert average} & 89.15 &  &  & 100.0 &  &  & 88.68 &  &  & 100.0 &  &  & 85.69 &  &  & 100.0 &  &  \\
    \midrule
    \multirow{5}{*}{\rotatebox[origin=c]{90}{\scriptsize Aggregation}} & Task Arithmetic~\cite{ilharco2023editing} & \textbf{61.62} & 61.21 & $-0.41$ & 33.0 & \textbf{41.4} & $+8.4$ & 48.64 & \textbf{48.69} & $+0.05$ & 10.1 & \textbf{13.2} & $+3.0$ & 43.17 & \textbf{43.21} & $+0.04$ & 10.7 & \textbf{11.3} & $+0.7$ \\
     & Fisher~\cite{matena2022merging} & 62.03 & \textbf{62.54} & $+0.51$ & 30.7 & \textbf{41.7} & $+11.0$ & 49.28 & \textbf{49.51} & $+0.23$ & 9.7 & \textbf{13.0} & $+3.3$ & 43.34 & \textbf{43.98} & $+0.63$ & 10.2 & \textbf{12.4} & $+2.2$ \\
     & TIES~\cite{yadav2023ties} & 64.26 & \textbf{64.41} & $+0.16$ & 40.8 & \textbf{56.3} & $+15.5$ & 49.21 & \textbf{49.51} & $+0.30$ & 11.0 & \textbf{15.8} & $+4.8$ & 44.01 & \textbf{44.05} & $+0.03$ & 11.5 & \textbf{13.4} & $+1.8$ \\
     & TSV-M~\cite{gargiulo2025tsv} & 74.79 & \textbf{76.54} & $+1.74$ & 62.8 & \textbf{77.3} & $+14.5$ & 62.94 & \textbf{63.94} & $+1.01$ & 31.1 & \textbf{40.3} & $+9.3$ & 53.96 & \textbf{54.30} & $+0.34$ & 22.0 & \textbf{23.9} & $+1.9$ \\
     & ISO-CTS~\cite{marczak2025isoc} & 73.63 & \textbf{74.90} & $+1.27$ & 56.6 & \textbf{71.7} & $+15.1$ & 64.94 & \textbf{66.48} & $+1.54$ & 31.5 & \textbf{39.2} & $+7.7$ & 56.30 & \textbf{56.93} & $+0.63$ & 22.2 & \textbf{25.6} & $+3.4$ \\
    \midrule
    \multirow{6}{*}{\rotatebox[origin=c]{90}{\scriptsize Reconstruction}} & RegMean~\cite{jin2023regmean} & 72.21 & \textbf{74.07} & $+1.86$ & 52.6 & \textbf{69.2} & $+16.6$ & 58.92 & \textbf{60.42} & $+1.49$ & 23.2 & \textbf{33.8} & $+10.7$ & 51.04 & \textbf{51.70} & $+0.66$ & 19.2 & \textbf{22.4} & $+3.2$ \\
     & RegMean++~\cite{nguyen2025regmeanpp} & 75.00 & \textbf{77.65} & $+2.65$ & 59.7 & \textbf{78.8} & $+19.2$ & 62.80 & \textbf{65.27} & $+2.47$ & 29.5 & \textbf{45.5} & $+16.1$ & 54.86 & \textbf{55.72} & $+0.86$ & 23.0 & \textbf{28.5} & $+5.5$ \\
     & LOT~\cite{sun2025lot} & 71.06 & \textbf{72.60} & $+1.54$ & 52.3 & \textbf{63.7} & $+11.4$ & 58.98 & \textbf{59.92} & $+0.94$ & 24.2 & \textbf{35.3} & $+11.1$ & 50.78 & \textbf{51.33} & $+0.55$ & 19.8 & \textbf{23.5} & $+3.7$ \\
     & WUDI~\cite{cheng2025wudi} & 76.33 & \textbf{78.27} & $+1.94$ & 66.5 & \textbf{79.1} & $+12.6$ & 63.77 & \textbf{65.00} & $+1.23$ & 33.3 & \textbf{42.2} & $+9.0$ & 54.90 & \textbf{55.21} & $+0.31$ & 22.7 & \textbf{26.0} & $+3.3$ \\
     & ACE-Merging~\cite{ace2025} & 75.46 & \textbf{78.33} & $+2.87$ & 63.8 & \textbf{76.9} & $+13.0$ & 60.27 & \textbf{63.54} & $+3.27$ & 30.0 & \textbf{51.2} & $+21.2$ & 50.95 & \textbf{52.30} & $+1.35$ & 18.5 & \textbf{23.8} & $+5.3$ \\
     & ReForge~\cite{li2026bmm} & 80.14 & \textbf{82.10} & $+1.96$ & 74.5 & \textbf{85.1} & $+10.6$ & 71.38 & \textbf{72.44} & $+1.06$ & 47.2 & \textbf{56.6} & $+9.4$ & 61.87 & \textbf{62.22} & $+0.35$ & 30.3 & \textbf{33.7} & $+3.4$ \\
    \midrule
    \multicolumn{2}{@{}l}{\textit{Average}} & 71.50 & \textbf{72.97} & $+1.46$ & 53.9 & \textbf{67.4} & $+13.4$ & 59.19 & \textbf{60.43} & $+1.23$ & 25.5 & \textbf{35.1} & $+9.6$ & 51.38 & \textbf{51.90} & $+0.52$ & 19.1 & \textbf{22.2} & $+3.1$ \\
    \bottomrule
  \end{tabular}%
  }
\end{table*}

\subsection{Experimental Setup}
\label{sec:setup}
We merge CLIP~\cite{radford2021clip} ViT-L/14 and ViT-B/32 experts on nested sets of 20, 50 and 100 tasks from CLIPMerge-100.
Each expert has a frozen zero-shot classification head. Details of CLIPMerge-100, search spaces, budgets, refinement rounds and random seeds are in App.~\ref{app:protocol}; compute cost and label requirements are in App.~\ref{app:cost}.

We evaluate \RaTW across eleven merging methods: five aggregation-based and six reconstruction-based.
For each method, we compare its original form (Native) with the variant incorporating \RaTW,
denoted by +\RaTW throughout. Models are selected by mean validation accuracy and evaluated on held-out test data.
For methods other than ReForge, +\RaTW uses grid search to jointly select the merging hyperparameters and the recovery exponent $\gamma$.
ReForge and \RFpp start from an ISO-CTS merged model.
Both select their hyperparameters by BO with 200 trials per round, and \RFpp additionally searches the recovery exponent $\gamma$.

\subsection{\RaTW across Merging Methods}
\label{sec:transfer}

Tables~\ref{tab:transfer} and~\ref{tab:transfer-b32} show that \RaTW improves tail20 in all 66 settings and Mean in 65. These gains span both aggregation-based and reconstruction-based methods, and extend to ReForge, the strongest Native method in both tables. The gains persist as the number of tasks increases but are smaller at 100 tasks, where \RaTW raises tail20 by 6.1 and 3.1 percentage points and Mean by 1.03 and 0.52 points on average on ViT-L/14 and ViT-B/32.

Across nearly all settings, Mean and tail20 improve together; the only exception is Task Arithmetic at 20 tasks on ViT-B/32, where Mean decreases slightly. These gains are not an artifact of test-set sampling: on ViT-B/32, a paired bootstrap over test images places the tail20 gain above zero with 95\% confidence in all 33 settings and the Mean gain in 28 (App.~\ref{app:ext-boot}). The gains concentrate on poorly retained tasks: retention on the other 80\% of tasks falls by 1.1 percentage points on average (App.~\ref{app:ext-head80}), a loss that the tail20 gains more than offset in Mean.

%% file: sec/4b_analysis.tex
\subsection{Understanding \RaTW}
\label{sec:analysis}
\raggedbottom

\paragraph{Separating the effects of task weighting and hyperparameter search.}
Because +\RaTW runs its own hyperparameter search, its gains could come from the new hyperparameters rather than from the task weights. Table~\ref{tab:controls}(a) separates the two by changing them one at a time, starting from Native's selected model. Task weights alone, with Native's hyperparameters, improve Mean by 0.45--1.97 percentage points and tail20 by 3.6--16.3 points in all four settings. +\RaTW's base hyperparameters alone, with uniform weights ($\gamma=0$), do not improve Mean; at 50 tasks they coincide with Native's. The gains therefore come from the task weights.

\begin{table*}[!t]
  \centering\footnotesize
  \renewcommand{\arraystretch}{0.95}
  \setlength{\tabcolsep}{3pt}
  \caption{\textbf{Analysis of \RaTW on ViT-B/32.} Each entry reports Mean / tail20. (a) Separate and combined effects of task weighting and hyperparameters, as gains over Native; ``Weights only'' uses the $\gamma$ selected by +\RaTW. (b) Transfer of recovery scores between ISO-CTS and RegMean, as gains over Native. (c) Comparison of weighting signals: Uniform reports absolute performance (\%), and the other rows report gains over Uniform. Panels (b) and (c) fix $\gamma=0.5$. (d) Comparison with other task-weighting methods, as gains over Native; AdaMerging is initialized at Native's coefficients. All gains are in percentage points.}
  \label{tab:controls}\label{tab:signals}
  \begin{tabular*}{0.47\textwidth}[t]{@{\extracolsep{\fill}}lccc@{}}
    \toprule
    \multicolumn{4}{@{}l}{\textbf{(a) Effects of weighting and hyperparameters}}\\
    Setting & Weights only & Hyperparam.\ only & Both (+\RaTW)\\
    \midrule
    ISO-CTS 20 & +0.45 / +3.6 & $-$1.15 / $-$4.7 & +1.27 / +15.1\\
    ISO-CTS 50 & +1.54 / +7.7 & 0 / 0 & +1.54 / +7.7\\
    RegMean 20 & +1.97 / +16.3 & $-$0.04 / +0.0 & +1.86 / +16.6\\
    RegMean 50 & +1.49 / +10.7 & 0 / 0 & +1.49 / +10.7\\
    \bottomrule
  \end{tabular*}\hfill
  \begin{tabular*}{0.49\textwidth}[t]{@{\extracolsep{\fill}}lcccc@{}}
    \toprule
    \multicolumn{5}{@{}l}{\textbf{(b) Recovery scores from another method}}\\
    Setting & Own $\beta$ & Other $\beta$ & Other ranking & $\rho$\\
    \midrule
    ISO-CTS 50 & +1.24 / +13.4 & +2.24 / +11.1 & +2.27 / +11.5 & 0.83\\
    ISO-CTS 100 & +0.63 / +3.4 & +1.13 / +2.5 & +1.17 / +3.1 & 0.91\\
    RegMean 50 & +1.00 / +5.1 & +0.36 / +5.2 & +0.39 / +4.8 & 0.83\\
    RegMean 100 & +0.56 / +1.5 & +0.32 / +2.0 & +0.38 / +1.8 & 0.91\\
    \bottomrule
  \end{tabular*}

  \vspace{4pt}
  \begin{tabular*}{\textwidth}{@{\extracolsep{\fill}}lcccccc@{}}
    \toprule
    \multicolumn{7}{@{}l}{\textbf{(c) Weighting signals}}\\
    & \multicolumn{3}{c}{ISO-CTS} & \multicolumn{3}{c}{RegMean}\\
    \cmidrule(lr){2-4}\cmidrule(l){5-7}
    Signal & $T=20$ & $T=50$ & $T=100$ & $T=20$ & $T=50$ & $T=100$\\
    \midrule
    Uniform & 73.63 / 56.6 & 64.94 / 31.5 & 56.30 / 22.2 & 72.21 / 52.6 & 58.92 / 23.2 & 51.04 / 19.2\\
    Recovery score $\beta$ & +0.45 / +3.6 & +1.24 / +13.4 & +0.63 / +3.4 & +1.23 / +8.0 & +1.00 / +5.1 & +0.56 / +1.5\\
    Accuracy gap & +0.45 / +3.4 & +1.38 / +12.8 & +0.55 / +2.9 & +1.15 / +7.9 & +0.91 / +4.1 & +0.41 / +1.4\\
    Reference error & +0.42 / +3.1 & +0.47 / +10.5 & $-$0.09 / +2.8 & +0.71 / +6.1 & +0.60 / +3.9 & +0.21 / +1.4\\
    KL (label-free) & +0.48 / +3.6 & +0.75 / +12.1 & $-$0.20 / +4.0 & +1.21 / +8.3 & +0.61 / +5.7 & $-$0.07 / +2.0\\
    Permutations of $\beta$, mean & $-$0.08 / +0.0 & $-$1.70 / $-$2.2 & $-$0.63 / $-$0.9 & $-$0.37 / $-$0.7 & $-$0.26 / $-$0.6 & $-$0.21 / $-$0.8\\
    \bottomrule
  \end{tabular*}
  \vspace{4pt}
  \begin{tabular*}{\textwidth}{@{\extracolsep{\fill}}lcccccc@{}}
    \toprule
    \multicolumn{7}{@{}l}{\textbf{(d) Other task-weighting methods}}\\
    & \multicolumn{3}{c}{ISO-CTS} & \multicolumn{3}{c}{TSV-M}\\
    \cmidrule(lr){2-4}\cmidrule(l){5-7}
    Method & $T=20$ & $T=50$ & $T=100$ & $T=20$ & $T=50$ & $T=100$\\
    \midrule
    MetaGPT & $-$0.53 / $-$0.5 & $-$7.32 / $-$8.1 & $-$5.41 / $-$3.2 & $-$3.57 / $-$8.2 & $-$6.73 / $-$10.7 & $-$6.81 / $-$7.9\\
    NAN & $-$0.38 / $-$1.0 & $-$5.49 / $-$8.7 & $-$1.07 / $-$1.9 & $-$2.02 / $-$10.2 & $-$0.06 / $-$1.1 & +0.52 / $-$1.4\\
    AdaMerging & +0.20 / 0.0 & +1.35 / +0.6 & +0.71 / $-$2.1 & +0.60 / +1.0 & +0.73 / $-$3.6 & +0.77 / $-$1.5\\
    +\RaTW & +1.27 / +15.1 & +1.54 / +7.7 & +0.63 / +3.4 & +1.74 / +14.5 & +1.01 / +9.3 & +0.34 / +1.9\\
    \bottomrule
  \end{tabular*}
\end{table*}

\paragraph{Sensitivity to the reference model.}
\RaTW measures recovery scores on a reference model, so its gains could depend on which model serves as the reference. Table~\ref{tab:controls}(b) tests this by swapping the scores of ISO-CTS and RegMean: each method is re-merged with the scores measured on the other method's merge, keeping its own Native hyperparameters and $\gamma=0.5$. The other method's scores (Other $\beta$) improve both Mean and tail20 over Native in all four settings. Keeping a method's own weight values but reassigning them to tasks according to the other method's $\beta$-ranking (Other ranking) gives similar gains. The two methods rank tasks similarly: the Spearman correlation between their rankings is 0.83 at 50 tasks and 0.91 at 100 tasks. Task weighting thus tolerates a change of reference model between these two merging methods.

\paragraph{Task weights and scoring signals.}
Table~\ref{tab:controls}(c) addresses two questions at Native's hyperparameters with $\gamma=0.5$: whether it matters which task receives which weight, and which signal should set the weights. For the first, we randomly permute the recovery weights across tasks. Averaged over ten permutations, neither Mean nor tail20 improves over uniform weights in any of the six settings, so the gains come from assigning large weights to the right tasks. For the second, $\beta$ is compared with three alternative signals. Reference error counts all mistakes of the reference model, whether or not the expert is correct; it yields smaller gains than $\beta$ in both metrics in all six settings, so it matters to count only the errors that the expert avoids. The accuracy gap between the expert and the reference yields smaller or equal gains than $\beta$ across all six settings (except Mean on ISO-CTS at $T=50$) and can be negative, requiring clipping, so we use $\beta$. The KL divergence between the expert's and the reference model's predictions requires no labels; it improves tail20 in all six settings but trails $\beta$ in Mean at 50 and 100 tasks, which makes it an alternative when labels are unavailable. All signals are defined in App.~\ref{app:signals}.

\paragraph{Comparison with other task-weighting methods.}
Table~\ref{tab:controls}(d) compares \RaTW with three other task-weighting methods on ISO-CTS and TSV-M. MetaGPT and NAN set the coefficients from parameter norms; they lower tail20 in all six settings and Mean in eleven of the twelve cases. AdaMerging, initialized at Native's coefficients, improves Mean in all six settings and exceeds \RaTW at 100 tasks by 0.08 and 0.43 points. Its tail20, however, changes by only $-3.6$ to $+1.0$ points, whereas \RaTW improves tail20 by 1.9--15.1 points in every setting (absolute results in App.~\ref{app:ext-learned}). MetaGPT, NAN and AdaMerging each set one coefficient per task vector and therefore do not apply to the six reconstruction-based methods, which \RaTW also improves.

\subsection{Evaluation of \RFpp}
\label{sec:main}

\paragraph{Task weighting.}
Table~\ref{tab:main} compares \RFpp with ReForge, the strongest Native method in Tables~\ref{tab:transfer}--\ref{tab:transfer-b32}. Three-round \RFpp (rows ``\RFpp, 3'') exceeds one-round ReForge (rows ``ReForge, 1'') in both Mean and tail20 at every task count, with \DFs and \DAs, on both backbones. This comparison is not budget-matched, however, since \RFpp uses three times as many BO trials. We therefore also run uniform-weight ReForge for three rounds with the same 600 trials (rows ``ReForge, 3''). \RFpp still improves both metrics in every setting. Averaged over the three task counts and over \DFs and \DAs, the gain of rows ``\RFpp, 3'' over rows ``ReForge, 3'' is 1.01 / 5.38 points in Mean / tail20 on ViT-L/14 and 1.15 / 6.22 on ViT-B/32 (computed from unrounded results). Thus, the advantage of \RFpp cannot be explained by the larger refinement budget alone.
\label{sec:budget}

\begin{figure*}[!t]
  \centering
  \includegraphics[width=0.78\textwidth]{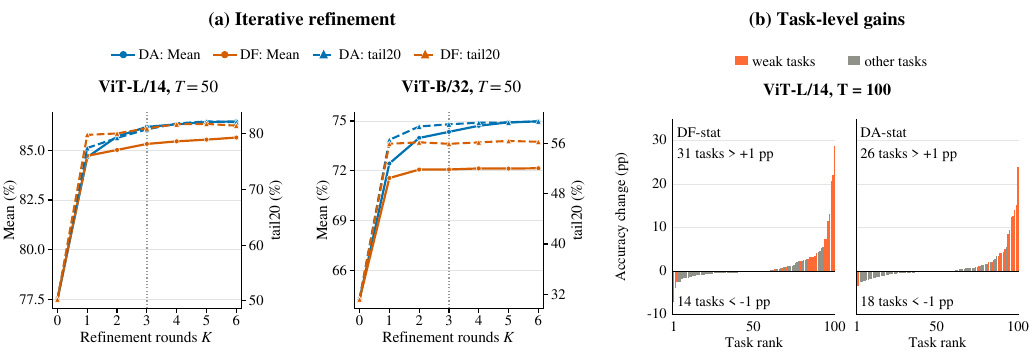}
  \vspace{-7pt}
  \caption{\textbf{Iterative refinement and task-level gains.} \textbf{(a)} Mean (solid) and tail20 (dashed) of \RFpp over refinement rounds. Round 0 is the initial ISO-CTS merge, each later round uses 200 BO trials, and the dotted line marks the default $K=3$. \textbf{(b)} Per-task accuracy differences between \RFpp and uniform-weight ReForge after three rounds, sorted in increasing order. Orange marks the weak tasks: the 20\% of tasks that the initial merge retains worst. Results use seed 0.}
  \label{fig:k-rounds}\label{fig:tasks}
  \vspace{-5pt}
\end{figure*}

\begin{table*}[!t]
  \centering\vspace{-3pt}
  \caption{\textbf{Evaluation of ReForge++ across refinement rounds.} Mean accuracy and tail20 (\%) on ViT-L/14 and ViT-B/32. ReForge uses uniform task weights and \RFpp uses recovery-aware task weights. Rows with the same number of rounds use the same budget of 200 BO trials per round, and all runs start from the same ISO-CTS merge. Bold marks the best result in each column.}
  \label{tab:main}
  \footnotesize
  \renewcommand{\arraystretch}{0.85}
  \setlength{\aboverulesep}{0.3ex}\setlength{\belowrulesep}{0.5ex}
  \setlength{\tabcolsep}{3.0pt}
  \begin{tabular}{@{}llc*{6}{r}@{\hspace{12pt}}*{6}{r}@{}}
    \toprule
    & & & \multicolumn{6}{c}{ViT-L/14} & \multicolumn{6}{c}{ViT-B/32} \\
    \cmidrule(lr){4-9}\cmidrule(l){10-15}
    & & & \multicolumn{2}{c}{$T=20$} & \multicolumn{2}{c}{$T=50$} & \multicolumn{2}{c}{$T=100$} & \multicolumn{2}{c}{$T=20$} & \multicolumn{2}{c}{$T=50$} & \multicolumn{2}{c}{$T=100$} \\
    \cmidrule(lr){4-5}\cmidrule(lr){6-7}\cmidrule(lr){8-9}\cmidrule(lr){10-11}\cmidrule(lr){12-13}\cmidrule(l){14-15}
    Statistics & Method & Rounds & Mean & tail20 & Mean & tail20 & Mean & tail20 & Mean & tail20 & Mean & tail20 & Mean & tail20 \\
    \midrule
    \multirow{4}{*}{\DFs} & ReForge & 1 & 89.96 & 89.3 & 83.29 & 70.4 & 72.45 & 38.6 & 80.10 & 74.2 & 70.15 & 46.5 & 61.19 & 29.3 \\
     & ReForge & 3 & 90.27 & 90.4 & 84.37 & 74.6 & 73.20 & 40.1 & 80.46 & 76.6 & 70.71 & 49.0 & 61.57 & 30.2 \\
     & \RFpp & 1 & 90.82 & 93.3 & 84.72 & 79.8 & 73.58 & 45.4 & 81.91 & 85.7 & 71.56 & 56.0 & 62.04 & 34.2 \\
     & \RFpp & 3 & 91.02 & 93.8 & 85.31 & \textbf{80.9} & 74.54 & 46.9 & 82.38 & \textbf{86.7} & 72.08 & 56.0 & 62.41 & 34.0 \\
    \midrule
    \multirow{4}{*}{\DAs} & ReForge & 1 & 90.02 & 88.6 & 83.69 & 69.7 & 73.36 & 39.3 & 80.14 & 74.5 & 71.38 & 47.2 & 61.87 & 30.3 \\
     & ReForge & 3 & 90.27 & 89.7 & 85.10 & 75.1 & 75.63 & 46.4 & 81.94 & 80.4 & 73.29 & 51.7 & 63.75 & 33.5 \\
     & \RFpp & 1 & 90.81 & 92.9 & 84.65 & 77.4 & 74.65 & 47.2 & 82.10 & 85.1 & 72.44 & 56.6 & 62.22 & 33.7 \\
     & \textbf{\RFpp} & 3 & \textbf{91.18} & \textbf{94.2} & \textbf{86.17} & 80.7 & \textbf{76.70} & \textbf{51.9} & \textbf{83.08} & 86.6 & \textbf{74.35} & \textbf{59.1} & \textbf{64.31} & \textbf{36.4} \\
    \bottomrule
  \end{tabular}
\vspace{-5pt}
\end{table*}

\paragraph{Iterative refinement.}
We next ask how much the additional rounds contribute. In Table~\ref{tab:main}, going from one to three rounds of \RFpp (rows ``\RFpp, 1'' and ``\RFpp, 3'') improves Mean in all twelve settings and tail20 in ten; the two exceptions, both with \DFs on ViT-B/32, change by at most 0.2 points. Figure~\ref{fig:k-rounds}(a) traces Mean and tail20 over six rounds at 50 tasks on both backbones. The first round gives by far the largest improvement, and later rounds add little, with occasional small decreases. Task weighting also acts mainly in the first round: afterwards, BO often selects $\gamma$ near zero, so later rounds refine the model with near-uniform weights (App.~\ref{app:fixed-beta}). Because each round costs another 200 BO trials, we use $K=3$ by default.

\paragraph{Which tasks benefit?}
Figure~\ref{fig:tasks}(b) shows which tasks account for the matched-budget gain on 100 ViT-L/14 tasks. It plots the per-task accuracy difference between three-round \RFpp and three-round uniform-weight ReForge, sorted in increasing order. The 20 tasks that the initial merge retained worst (orange) account for 8 and 9 of the ten largest gains under \DFs and \DAs, and improve by 6.53 and 6.10 percentage points on average. The effect is uneven across the remaining tasks: across all 100 tasks, 31 and 26 gain more than one point, whereas 14 and 18 lose more than one point. Thus, at a matched budget, task weighting particularly benefits the tasks that the initial merge retained worst.

\input{sec/4_vlm}
\FloatBarrier

%% file: sec/4_vlm.tex
\subsection{Extension to Vision-Language Models}
\label{sec:vlm}
\input{sec/tab_vlm_compact}

The experiments above merge fully fine-tuned CLIP classifiers. This section tests whether \RaTW extends to generative vision-language models with LoRA experts. On InternVL2.5-1B~\cite{chen2024internvl25} and Qwen2-VL-7B~\cite{wang2024qwen2vl}, we merge ten task-specific LoRA experts and add \RaTW to five methods: Task Arithmetic~\cite{ilharco2023editing}, TIES~\cite{yadav2023ties}, both ReForge variants~\cite{li2026bmm}, and OptMerge~\cite{wei2026optmerge}, which is designed for multimodal large language models. In these tasks, an answer can receive partial credit, i.e., a score between 0 and 1, so we cannot simply count the samples that the expert gets right and the merged model gets wrong. Instead, $k_t$ adds up how much higher the expert scores than the merged model on each sample where it scores higher; when all scores are 0 or 1, this equals the count used in Eq.~\eqref{eq:beta}. Table~\ref{tab:vlm-main} shows that \RaTW improves both Mean and tail20 in all ten comparisons, and, as on CLIP, the gains are larger in tail20 (e.g., from 72.2 to 85.7 for Task Arithmetic on Qwen2-VL-7B). Three-round \RFpp with \DAs achieves the best results on both backbones. Setup and full results are in App.~\ref{app:vlm}.

%% file: sec/tab_vlm_compact.tex
\begin{table}[!t]
\centering\footnotesize
\newcommand{\vlmc}[2]{#1\,/\,#2}
\caption{\textbf{\RaTW on vision-language models.} Mean / tail20 (\%) on ten tasks, where tail20 averages the two worst-retained tasks.}\vspace{-3pt}
\label{tab:vlm-main}
\setlength{\tabcolsep}{2.5pt}
\resizebox{\columnwidth}{!}{%
\begin{tabular}{@{}lcc@{\hspace{8pt}}cc@{}}
\toprule
& \multicolumn{2}{c}{InternVL2.5-1B} & \multicolumn{2}{c}{Qwen2-VL-7B}\\
\cmidrule(lr){2-3}\cmidrule(l){4-5}
Method & Native & +\RaTW & Native & +\RaTW\\
\midrule
Task Arithmetic & \vlmc{68.93}{66.2} & \textbf{\vlmc{70.62}{74.7}} & \vlmc{82.56}{72.2} & \textbf{\vlmc{83.63}{85.7}}\\
TIES & \vlmc{70.31}{70.7} & \textbf{\vlmc{70.48}{71.7}} & \vlmc{84.94}{88.4} & \textbf{\vlmc{85.18}{89.8}}\\
OptMerge & \vlmc{67.74}{63.1} & \textbf{\vlmc{68.53}{68.8}} & \vlmc{85.16}{90.9} & \textbf{\vlmc{85.66}{92.9}}\\
ReForge-DF & \vlmc{71.50}{72.6} & \textbf{\vlmc{71.82}{73.4}} & \vlmc{85.50}{90.6} & \textbf{\vlmc{85.99}{93.3}}\\
ReForge-DA & \vlmc{74.05}{78.3} & \textbf{\vlmc{74.95}{83.8}} & \vlmc{87.10}{96.2} & \textbf{\vlmc{87.30}{97.7}}\\
\midrule
\RFpp-DF & -- & \vlmc{71.39}{72.9} & -- & \vlmc{86.99}{93.6}\\
\RFpp-DA & -- & \textbf{\vlmc{77.31}{91.8}} & -- & \textbf{\vlmc{87.67}{99.1}}\\
\bottomrule
\end{tabular}}
\vspace{-3pt}
\end{table}

%% file: sec/5_conclusion.tex
\section{Conclusion and Future Work}
\label{sec:concl}

We introduce Recovery-aware Task Weighting (\RaTW), a plug-and-play method that assigns larger weights to tasks whose expert capabilities the merged model retains poorly. To evaluate merging at large scale, we construct CLIPMerge-100, a 100-task benchmark with task-specific experts on two CLIP backbones. Across eleven merging methods with 20, 50 and 100 tasks, \RaTW improves both mean accuracy and tail20 in 65 of 66 settings, and it also improves both metrics for five methods on two vision-language models. We further develop \RFpp, which extends ReForge with recovery-aware weighting and iterative refinement and outperforms all evaluated baselines, including a matched-budget control.

Future work will explore more efficient search and refinement strategies to reduce the Bayesian optimization cost of \RFpp. We also plan to study stronger label-free recovery signals and selection criteria that fully remove the need for labeled validation data.

%% file: sec/X_appendix.tex
\renewcommand{\topfraction}{0.95}
\renewcommand{\dbltopfraction}{0.95}
\renewcommand{\textfraction}{0.05}
\renewcommand{\floatpagefraction}{0.75}
\renewcommand{\dblfloatpagefraction}{0.75}
\setcounter{topnumber}{3}
\setcounter{dbltopnumber}{3}
\setcounter{totalnumber}{5}
\raggedbottom
\makeatletter
\setlength{\@fptop}{0pt}\setlength{\@fpbot}{0pt plus 1fil}
\setlength{\@dblfptop}{0pt}\setlength{\@dblfpbot}{0pt plus 1fil}
\setlength{\@fpsep}{14pt}\setlength{\@dblfpsep}{14pt}
\makeatother
\input{sec/X_appendix_B}

\section{Experimental Setup and CLIPMerge-100}
\label{app:protocol}

\subsection{Tasks, experts and data splits}
Both CLIP backbones use the nested task lists in Table~\ref{tab:app-provenance}, with the first 20 tasks following TALL-20 in FusionBench~\cite{tang2024fusionbench}. Each expert is trained with Adam, a learning rate of $10^{-5}$, no weight decay, a 4{,}000-step cosine schedule, batch size 128 and seed 42. Only the vision encoder is trained, without augmentation; the text encoder, projection and logit scale remain frozen. The 20 ViT-B/32 FusionBench experts are retrained with our validation images excluded. Experts and merged models share frozen task-specific CLIP zero-shot heads.

Both backbones share the data splits in Table~\ref{tab:app-provenance}. Validation uses published splits for 17 tasks, held-out toy instances for the two smallNORB tasks, and a stratified 10\% of the training split for the remaining 81 tasks (seed 0). For ten of the latter tasks, we also construct training/test splits: seven single-split sources use an 80/20 split by a fixed hash of (label, key), while mit\_indoor67, pokemon and human\_action use image-hash splits as specified in Table~\ref{tab:app-provenance}. A fixed subset of at most 1{,}000 validation images per task (seed 0) serves both recovery-score estimation and model selection. \DAs uses 128 training images per task (seed 0), shared across methods and rounds.

Grouping is verified only for smallNORB, whose toy-instance IDs were recovered by exact pixel matching for all 24{,}300 source training images. Source reports describe slide-disjoint PCam and video-disjoint KITTI splits, but per-row group IDs were not retained; patient grouping for eeg\_seizure\_spectrogram is undocumented. Grouping is neither imposed nor checked for other tasks, so validation or test images may share subjects with training images. Within each task, exact cross-split duplicates are removed using SHA-1 of stored image bytes, or pixel arrays for MedMNIST: test images take precedence over training and validation, and validation images over training. Within-split duplicates and near duplicates remain; near duplicates are only counted.

A cross-task check over all 5.76M images identified about 11k test images matching another task's training or validation data, using exact matching and near-duplicate screening with dHash Hamming distance $\leq2$ and thumbnail verification. Near-duplicate matches were not manually verified. Eight tasks exceed a 5\% test-image match rate---tomato\_leaf, plantvillage, caltech101, mit\_indoor67, bangla\_vs\_assamese\_script, imagewoof, sun397 and stanford\_dogs---while all others are at about 1\% or below. Across all 66 settings, excluding these eight tasks from evaluation while retaining their experts in the merge changes the average \RaTW gains within each backbone and task count by at most 0.2\,pp in Mean and 0.3\,pp in tail20. After exclusion, tail20 improves in all 66 settings and Mean in 61, compared with 65 before exclusion.

\input{sec/tab_tasks}

\paragraph{Software and hardware.}
\label{app:repro}
The CLIP environment uses PyTorch 2.7.1 (CUDA 12.6), FusionBench 0.2.33, Transformers 4.55.4 and Optuna 4.8.0. Experiments use RTX 6000 Ada (48\,GB), A100 and L40S GPUs; all timings in Sec.~\ref{app:cost} were measured on RTX 6000 Ada.

\subsection{Model selection and search spaces}
\label{app:search}
Models are selected by mean validation accuracy with equal weight per task; only selected models are evaluated on test. Table~\ref{tab:app-native-grids} lists the Native search grids for the ten non-ReForge methods. For a grid of $c$ configurations, +\RaTW searches its Cartesian product with $\gamma\in\{0,0.25,0.5,1,2,3\}$, giving $6c$ configurations: $c$ reused Native configurations at $\gamma=0$ and $5c$ new configurations. Ties in mean validation accuracy favor smaller $\gamma$, followed by a fixed order of the Native configurations---the lexicographic order of their canonical JSON serialization. Candidate counts do not measure total computational cost, which also includes recovery scoring, statistics collection and final test evaluation; measured costs are reported in Sec.~\ref{app:cost}.

\begin{table}[!htbp]
\centering\footnotesize
\setlength{\tabcolsep}{3pt}
\caption{\textbf{Initial refinement model $\theta^{(0)}$.} Entries report Mean / tail20 (\%) for the separately selected ISO-CTS anchor. The Native ISO-CTS baselines in Tables~\ref{tab:transfer} and~\ref{tab:transfer-b32} use the search grid in Table~\ref{tab:app-native-grids} and need not coincide with this initialization. Native ISO-CTS matches or exceeds the anchor in both metrics at $T=20$ and $T=50$ on both backbones, and at $T=100$ on ViT-B/32.}
\label{tab:app-baselines}
\begin{tabular}{@{}lccc@{}}
\toprule
Backbone & $T=20$ & $T=50$ & $T=100$\\
\midrule
ViT-L/14 & 87.74 / 82.7 & 77.51 / 50.3 & 68.16 / 31.8\\
ViT-B/32 & 73.63 / 56.6 & 64.24 / 31.2 & 56.30 / 22.2\\
\bottomrule
\end{tabular}
\end{table}

Refinement starts from a separately selected ISO-CTS model $\theta^{(0)}$, reported in Table~\ref{tab:app-baselines}. This initialization is selected independently of the cross-method Native ISO-CTS baseline, which uses the 57-configuration grid in Table~\ref{tab:app-native-grids}, so the two models need not coincide. In round $r$, \RFpp uses its preceding model $\theta^{(r-1)}$ as both the recovery reference and the anchor, whereas uniform-weight ReForge uses its own preceding model as the anchor and requires no recovery reference. Each round selects a new candidate; the preceding model itself is not included in the candidate pool.

\input{sec/app_native_grids}

ReForge and \RFpp share the reconstruction search space in Table~\ref{tab:app-space}; only weighted variants additionally search $\gamma$. BO uses ReForge's Gaussian-process sampler with expected improvement in Optuna~\cite{akiba2019optuna}. Unless stated otherwise, the BO seed is 0 and the budget is 200 trials per round. Replications with other BO seeds are reported in Sec.~\ref{app:ext-seeds}.

\begin{table}[!htbp]
  \centering\footnotesize
  \setlength{\tabcolsep}{4pt}
  \caption{\textbf{ReForge reconstruction search space.} Both backbones use the same 15 reconstruction variables and ranges. ViT-B/32 divides encoder layers into segments 0--3, 4--7 and 8--11, while ViT-L/14 uses 0--7, 8--15 and 16--23; layer indices are zero-based. Weighted variants additionally search $\gamma\in[0,3]$ on ViT-B/32 and $\gamma\in[0,2]$ on ViT-L/14.}
  \label{tab:app-space}
  \begin{tabular}{@{}llcll@{}}
    \toprule
    Variable & Per & \# & Range & Scale \\
    \midrule
    Task-vector scale & segment & 3 & $[0,2]$ & linear \\
    $\lambda$, stacked q/k/v projection & segment & 3 & $[10^{-4},10^{2}]$ & log \\
    $\lambda$, attention output & segment & 3 & $[10^{-4},10^{2}]$ & log \\
    $\lambda$, MLP fc1 & segment & 3 & $[10^{-4},10^{2}]$ & log \\
    $\lambda$, MLP fc2 & segment & 3 & $[10^{-4},10^{2}]$ & log \\
    \bottomrule
  \end{tabular}
\end{table}

\section{Additional \RaTW Results}
\label{app:extended}
This section extends the cross-method results of Sec.~\ref{sec:transfer} and the controls of Sec.~\ref{sec:analysis}.

\subsection{Head and tail retention}
\label{app:ext-head80}
Table~\ref{tab:app-head80} examines the trade-off between head and tail retention by comparing Native and +\RaTW on task sets fixed by Native's test retention. head80, the mean retention over the 80\% of tasks outside Native's tail20 set, decreases in 60 of the 66 settings. Averaged over all 66 settings, it changes by $-1.0$\,pp; the largest decrease is 5.6\,pp, for Fisher on ViT-B/32 at 20 tasks. Meanwhile, retention on Native's fixed tail20 set increases by 3.7--16.3\,pp when averaged over the eleven methods within each backbone and task count. These settings share experts, data and methods across nested task sets and therefore are not independent replicates.

\input{sec/tab_head80}

\subsection{Bootstrap confidence intervals}
\label{app:ext-boot}
Table~\ref{tab:app-boot} reports 95\% paired bootstrap intervals for the ViT-B/32 gains in Table~\ref{tab:transfer-b32}. We perform 10{,}000 bootstrap resamples within each task, using the same resampled test-image indices for Native, +\RaTW, and the expert. Each model's tail20 set is recomputed in every resample. The tail20 intervals lie above zero in all 33 settings. The Mean intervals lie above zero in 28 settings, contain zero in four (TIES at 20 and 100 tasks, and Task Arithmetic at 50 and 100 tasks), and lie below zero for Task Arithmetic at 20 tasks.

Table~\ref{tab:app-boot-l14} applies the same protocol to all 33 ViT-L/14 settings in Table~\ref{tab:transfer}. For three settings whose selected checkpoints had been deleted (TIES, TSV-M and ACE-Merging at 20 tasks), the selected model was rebuilt from its stored hyperparameters and recovery exponent; each rebuilt model reproduces the stored validation mean and per-task test accuracies exactly. The tail20 intervals lie above zero in all 33 settings, while the Mean intervals lie above zero in 32 and contain zero for TIES at 100 tasks. These intervals quantify uncertainty from test-sample resampling only; they do not account for model selection or search randomness. Results across search seeds are reported separately in Table~\ref{tab:app-seeds}.

\begin{table*}[!tp]
\centering\footnotesize
\setlength{\tabcolsep}{3pt}
\caption{\textbf{Paired bootstrap intervals for \RaTW gains on ViT-B/32.} Gains are +\RaTW minus Native, in percentage points. Brackets contain 95\% percentile intervals from 10{,}000 resamples of test images within each task. Native, +\RaTW and the expert use the same resampled indices, and each model's tail20 set is recomputed in every resample. Point estimates match the $\Delta$ columns of Table~\ref{tab:transfer-b32}. Italics indicate intervals that contain zero.}
\label{tab:app-boot}
\begin{tabular}{@{}lcccccc@{}}
\toprule
 & \multicolumn{2}{c}{$T=20$} & \multicolumn{2}{c}{$T=50$} & \multicolumn{2}{c}{$T=100$}\\
\cmidrule(lr){2-3}\cmidrule(lr){4-5}\cmidrule(l){6-7}
Method & $\Delta$Mean & $\Delta$tail20 & $\Delta$Mean & $\Delta$tail20 & $\Delta$Mean & $\Delta$tail20\\
\midrule
Task Arithmetic & $-$0.41 {\scriptsize[$-$0.56, $-$0.26]} & +8.4 {\scriptsize[+7.8, +9.0]} & \textit{+0.05} {\scriptsize\textit{[$-$0.09, +0.19]}} & +3.0 {\scriptsize[+2.8, +3.3]} & \textit{+0.04} {\scriptsize\textit{[$-$0.04, +0.12]}} & +0.7 {\scriptsize[+0.5, +0.8]}\\
Fisher & +0.51 {\scriptsize[+0.31, +0.71]} & +11.0 {\scriptsize[+10.0, +12.0]} & +0.23 {\scriptsize[+0.08, +0.38]} & +3.3 {\scriptsize[+3.1, +3.6]} & +0.63 {\scriptsize[+0.51, +0.75]} & +2.2 {\scriptsize[+2.0, +2.4]}\\
TIES & \textit{+0.16} {\scriptsize\textit{[$-$0.02, +0.33]}} & +15.5 {\scriptsize[+14.9, +16.1]} & +0.30 {\scriptsize[+0.11, +0.49]} & +4.8 {\scriptsize[+4.4, +5.1]} & \textit{+0.03} {\scriptsize\textit{[$-$0.07, +0.13]}} & +1.8 {\scriptsize[+1.7, +2.0]}\\
TSV-M & +1.74 {\scriptsize[+1.60, +1.89]} & +14.5 {\scriptsize[+13.8, +15.1]} & +1.01 {\scriptsize[+0.86, +1.15]} & +9.3 {\scriptsize[+8.9, +9.7]} & +0.34 {\scriptsize[+0.23, +0.45]} & +1.9 {\scriptsize[+1.6, +2.4]}\\
ISO-CTS & +1.27 {\scriptsize[+1.07, +1.46]} & +15.1 {\scriptsize[+14.6, +15.7]} & +1.54 {\scriptsize[+1.41, +1.68]} & +7.7 {\scriptsize[+7.3, +8.2]} & +0.63 {\scriptsize[+0.49, +0.77]} & +3.4 {\scriptsize[+3.2, +3.6]}\\
\midrule
RegMean & +1.86 {\scriptsize[+1.72, +1.99]} & +16.6 {\scriptsize[+16.1, +17.1]} & +1.49 {\scriptsize[+1.35, +1.64]} & +10.7 {\scriptsize[+10.2, +11.0]} & +0.66 {\scriptsize[+0.53, +0.78]} & +3.2 {\scriptsize[+2.9, +3.5]}\\
RegMean++ & +2.65 {\scriptsize[+2.51, +2.80]} & +19.2 {\scriptsize[+18.4, +19.7]} & +2.47 {\scriptsize[+2.31, +2.63]} & +16.1 {\scriptsize[+15.5, +16.6]} & +0.86 {\scriptsize[+0.72, +1.00]} & +5.5 {\scriptsize[+5.0, +5.9]}\\
LOT & +1.54 {\scriptsize[+1.42, +1.67]} & +11.4 {\scriptsize[+10.9, +11.8]} & +0.94 {\scriptsize[+0.78, +1.10]} & +11.1 {\scriptsize[+10.5, +11.6]} & +0.55 {\scriptsize[+0.41, +0.69]} & +3.7 {\scriptsize[+3.3, +4.1]}\\
WUDI & +1.94 {\scriptsize[+1.81, +2.08]} & +12.6 {\scriptsize[+12.2, +13.1]} & +1.23 {\scriptsize[+1.10, +1.36]} & +9.0 {\scriptsize[+8.5, +9.3]} & +0.31 {\scriptsize[+0.19, +0.42]} & +3.3 {\scriptsize[+2.9, +3.5]}\\
ACE & +2.87 {\scriptsize[+2.73, +3.00]} & +13.0 {\scriptsize[+12.5, +13.6]} & +3.27 {\scriptsize[+3.04, +3.49]} & +21.2 {\scriptsize[+20.5, +21.8]} & +1.35 {\scriptsize[+1.22, +1.48]} & +5.3 {\scriptsize[+4.8, +5.7]}\\
ReForge & +1.96 {\scriptsize[+1.83, +2.09]} & +10.6 {\scriptsize[+10.2, +11.0]} & +1.06 {\scriptsize[+0.90, +1.22]} & +9.4 {\scriptsize[+8.8, +10.0]} & +0.35 {\scriptsize[+0.23, +0.47]} & +3.4 {\scriptsize[+3.1, +3.7]}\\
\bottomrule
\end{tabular}
\end{table*}

\begin{table*}[!tp]
\centering\footnotesize
\setlength{\tabcolsep}{3pt}
\caption{\textbf{Paired bootstrap intervals for \RaTW gains on ViT-L/14.} The bootstrap protocol, units and notation follow Table~\ref{tab:app-boot}. Point estimates match the $\Delta$ columns of Table~\ref{tab:transfer}, except Task Arithmetic at $T=20$: its stored test report gives a Mean of 76.06 ($+2.77$), whereas Table~\ref{tab:transfer} lists 76.04 ($+2.74$); the interval uses the stored report.}
\label{tab:app-boot-l14}
\begin{tabular}{@{}lcccccc@{}}
\toprule
 & \multicolumn{2}{c}{$T=20$} & \multicolumn{2}{c}{$T=50$} & \multicolumn{2}{c}{$T=100$}\\
\cmidrule(lr){2-3}\cmidrule(lr){4-5}\cmidrule(l){6-7}
Method & $\Delta$Mean & $\Delta$tail20 & $\Delta$Mean & $\Delta$tail20 & $\Delta$Mean & $\Delta$tail20\\
\midrule
Task Arithmetic & +2.77 {\scriptsize[+2.63, +2.91]} & +22.9 {\scriptsize[+22.3, +23.5]} & +0.77 {\scriptsize[+0.63, +0.92]} & +6.9 {\scriptsize[+6.5, +7.2]} & +0.27 {\scriptsize[+0.15, +0.39]} & +2.6 {\scriptsize[+2.3, +2.8]}\\
Fisher & +1.27 {\scriptsize[+1.12, +1.42]} & +14.0 {\scriptsize[+13.2, +14.8]} & +0.86 {\scriptsize[+0.70, +1.03]} & +2.7 {\scriptsize[+2.4, +3.0]} & +1.35 {\scriptsize[+1.20, +1.50]} & +2.7 {\scriptsize[+2.4, +3.0]}\\
TIES & +3.71 {\scriptsize[+3.58, +3.84]} & +24.8 {\scriptsize[+24.2, +25.5]} & +0.52 {\scriptsize[+0.42, +0.62]} & +2.7 {\scriptsize[+2.5, +2.9]} & \textit{+0.02} {\scriptsize\textit{[$-$0.04, +0.09]}} & +0.4 {\scriptsize[+0.3, +0.6]}\\
TSV-M & +2.04 {\scriptsize[+1.95, +2.14]} & +10.6 {\scriptsize[+10.2, +11.1]} & +2.37 {\scriptsize[+2.19, +2.54]} & +19.0 {\scriptsize[+18.2, +19.8]} & +0.80 {\scriptsize[+0.69, +0.91]} & +4.8 {\scriptsize[+4.5, +5.2]}\\
ISO-CTS & +1.42 {\scriptsize[+1.30, +1.53]} & +4.9 {\scriptsize[+4.6, +5.3]} & +3.01 {\scriptsize[+2.84, +3.19]} & +18.4 {\scriptsize[+17.5, +19.3]} & +2.13 {\scriptsize[+2.00, +2.26]} & +6.8 {\scriptsize[+6.3, +7.3]}\\
\midrule
RegMean & +3.00 {\scriptsize[+2.88, +3.12]} & +19.7 {\scriptsize[+19.2, +20.1]} & +3.16 {\scriptsize[+3.00, +3.33]} & +19.8 {\scriptsize[+18.6, +20.4]} & +0.78 {\scriptsize[+0.64, +0.92]} & +8.2 {\scriptsize[+7.6, +8.5]}\\
RegMean++ & +2.88 {\scriptsize[+2.78, +2.99]} & +15.7 {\scriptsize[+15.3, +16.1]} & +3.07 {\scriptsize[+2.90, +3.25]} & +19.4 {\scriptsize[+18.5, +20.3]} & +0.82 {\scriptsize[+0.67, +0.97]} & +7.9 {\scriptsize[+7.4, +8.5]}\\
LOT & +2.83 {\scriptsize[+2.70, +2.96]} & +20.1 {\scriptsize[+19.4, +20.6]} & +2.56 {\scriptsize[+2.39, +2.74]} & +18.3 {\scriptsize[+17.4, +19.3]} & +0.88 {\scriptsize[+0.76, +0.99]} & +5.3 {\scriptsize[+5.0, +5.8]}\\
WUDI & +1.65 {\scriptsize[+1.56, +1.74]} & +9.6 {\scriptsize[+9.2, +9.9]} & +3.51 {\scriptsize[+3.33, +3.70]} & +25.2 {\scriptsize[+24.3, +25.9]} & +1.17 {\scriptsize[+1.04, +1.30]} & +7.0 {\scriptsize[+6.5, +7.5]}\\
ACE & +1.94 {\scriptsize[+1.82, +2.07]} & +12.0 {\scriptsize[+11.4, +12.5]} & +3.94 {\scriptsize[+3.77, +4.11]} & +19.7 {\scriptsize[+19.0, +20.5]} & +1.78 {\scriptsize[+1.63, +1.94]} & +13.2 {\scriptsize[+12.6, +13.9]}\\
ReForge & +0.78 {\scriptsize[+0.69, +0.87]} & +4.3 {\scriptsize[+4.0, +4.6]} & +0.96 {\scriptsize[+0.82, +1.09]} & +7.7 {\scriptsize[+6.9, +8.3]} & +1.29 {\scriptsize[+1.17, +1.43]} & +7.9 {\scriptsize[+7.3, +8.5]}\\
\bottomrule
\end{tabular}
\end{table*}

\subsection{Why gains shrink at 100 tasks}
\label{app:ext-scale}
Table~\ref{tab:app-scale} examines the weights assigned to weak tasks and their corresponding test gains across task counts, with weak-task membership fixed by Native's tail20 set. Tail lift, the weight share of these tasks relative to uniform weighting, decreases from approximately 2.1--2.4 at 20 and 50 tasks to approximately 1.9--2.0 at 100 tasks. At 100 tasks, the selected exponent $\gamma$ averages 1.0--1.1, below the search-grid maximum of 3. Larger task weights nevertheless remain associated with larger per-task test gains, with an average Spearman correlation of approximately 0.7.

The Response measure divides the average tail-task accuracy gain by the extra tail lift (tail lift minus one) within each setting, then averages these ratios across settings. From 20 to 100 tasks, Response decreases from 14.9 to 8.3\,pp on ViT-L/14 and from 16.7 to 4.9\,pp on ViT-B/32. The decrease is larger on ViT-B/32; together, these observations associate smaller gains at 100 tasks with both less extra weight on weak tasks and lower gains per unit of extra weight. These correlational comparisons do not establish the causal contribution of either factor.

\begin{table}[t]
\centering
\caption{\textbf{Weight concentration and response across task counts.} $n_q$ and $n_g$ count settings with stored task weights and per-task test accuracies, respectively. Tail and head membership is fixed by Native's test retention. Tail lift is the weight share of Native's tail20 tasks relative to their share under uniform weights, using the selected $\gamma$; $\rho$ is the Spearman correlation between $q_t$ and per-task test gain. Tail and Head report test-accuracy changes from Native in percentage points, averaged over the fixed tail20 set and its complement. Response is computed as Tail$/(\text{Tail lift}-1)$ within each setting and then averaged across settings; it is not the ratio of the column means. $\bar\gamma$, Tail lift, $\rho$ and Response are averaged over the $n_q$ settings, while Tail and Head are averaged over the $n_g$ settings. Every row covers the ten non-ReForge methods; ReForge is excluded, and ISO-CTS uses the extended-grid runs.}
\label{tab:app-scale}
\footnotesize
\setlength{\tabcolsep}{3pt}
\begin{tabular}{@{}llccccccc@{}}
\toprule
 & $T$ & $n_q$/$n_g$ & $\bar\gamma$ & Tail lift & $\rho$ & Tail & Head & Response\\
\midrule
\multirow{3}{*}{L/14} & 20  & 10/10 & 1.0 & 2.33 & 0.74 & $+15.9$ & $-1.04$ & 14.9\\
                      & 50  & 10/10 & 1.0 & 2.40 & 0.74 & $+15.9$ & $-1.01$ & 12.7\\
                      & 100 & 10/10 & 1.0 & 1.95 & 0.68 & $+6.6$  & $-0.41$ & 8.3\\
\midrule
\multirow{3}{*}{B/32} & 20  & 10/10 & 1.0 & 2.15 & 0.68 & $+16.0$ & $-2.23$ & 16.7\\
                      & 50  & 10/10 & 1.3 & 2.29 & 0.73 & $+10.4$ & $-1.03$ & 11.7\\
                      & 100 & 10/10 & 1.1 & 1.91 & 0.70 & $+3.4$  & $-0.17$ & 4.9\\
\bottomrule
\end{tabular}
\end{table}

\subsection{Definitions of weighting signals}
\label{app:signals}
Table~\ref{tab:controls}(c) compares scoring signals at Native's selected hyperparameters, using the same weight mapping in Eq.~\eqref{eq:q} with $\gamma=0.5$. For $n=|\mathcal C_t|$, let $k_t$ count expert-correct/reference-wrong samples and $h_t$ count expert-wrong/reference-correct samples, so that
\begin{equation}
 \operatorname{Acc}(E_t)-\operatorname{Acc}(R)=(k_t-h_t)/n.
 \label{eq:app-gap}
\end{equation}
The recovery signal, $\beta_t=\max(k_t,1)/n$, counts only errors of the reference model that the expert avoids. The reference-error signal, $e_t=\max(\#\{R\text{ wrong}\},1)/n$, counts all reference errors, regardless of whether the expert is correct. The accuracy-gap signal, $\max(k_t-h_t,1)/n$, subtracts cases where the reference is correct but the expert is wrong before applying the count floor of one. This floor applies to at most two tasks per setting for the accuracy-gap signal and is inactive for recovery in these settings because every task has $k_t\geq1$.

The label-free signal is
\[
s_t=\frac{1}{n}\sum_{x\in\mathcal C_t}\mathrm{KL}\bigl(p_{E_t}(\cdot\mid x)\,\|\,p_R(\cdot\mid x)\bigr),
\]
which measures KL divergence from the expert to the reference on the same validation images used for $\beta$. The probabilities are computed over task $t$'s classes using softmax at CLIP's own logit scale and floored at $10^{-12}$; no labels are used. The signal's definition, direction, and evaluated settings were fixed before any KL values were computed, but after the other signals' results were known.

The permutation control randomly reassigns the $\beta$-derived weight vector $\mathbf q$ across tasks, preserving the weight values while testing whether their assignment to tasks matters. We use ten permutations generated with NumPy's \texttt{default\_rng} and seeds 1--10, fixed before any run, and report the mean performance gains of the resulting merges.

\subsection{Other task-weighting methods}
\label{app:ext-learned}
We compare \RaTW with three alternatives for setting task coefficients: the training-free rules MetaGPT~\cite{zhou2024metagpt} and NAN~\cite{si2025nan}, entropy-based coefficient learning as in AdaMerging~\cite{yang2024adamerging}, and supervised coefficient learning using the same labeled validation data as \RaTW. Table~\ref{tab:app-coef-iso} reports absolute results for ISO-CTS and TSV-M on ViT-B/32; Table~\ref{tab:controls}(d) reports the corresponding gains over Native. For MetaGPT and NAN, the output scale is re-selected on validation. NAN is applied to task vectors: coefficients computed from full fine-tuned parameters differ from uniform by at most 0.02\%. AdaMerging starts from Native's selected hyperparameters with equal task coefficients and runs 500 Adam steps at a learning rate of 0.001 to minimize prediction entropy on unlabeled validation inputs; we report the last iterate.

MetaGPT and NAN reduce tail20 relative to Native in all six settings, whereas AdaMerging improves Mean in all six. At 100 tasks, AdaMerging exceeds +\RaTW in Mean for both merging methods, but its tail20 remains below +\RaTW in every setting.

\begin{table}[!tp]
  \centering\footnotesize
  \caption{\textbf{Alternative task-weighting methods on ViT-B/32.} The table compares task-coefficient schemes for ISO-CTS and TSV-M at 20, 50 and 100 tasks. Entries report test Mean / tail20 (\%). Native and +\RaTW results repeat Table~\ref{tab:transfer-b32}. NAN is applied to task vectors, as described in Sec.~\ref{app:ext-learned}.}
  \label{tab:app-coef-iso}
  {\setlength{\tabcolsep}{4pt}
\begin{tabular}{@{}llccc@{}}
\toprule
Method & Coefficients & $T=20$ & $T=50$ & $T=100$\\
\midrule
ISO-CTS & Native (uniform) & 73.63 / 56.6 & 64.94 / 31.5 & 56.30 / 22.2\\
 & MetaGPT & 73.10 / 56.1 & 57.62 / 23.4 & 50.89 / 19.0\\
 & NAN & 73.24 / 55.6 & 59.45 / 22.8 & 55.24 / 20.2\\
 & AdaMerging & 73.82 / 56.6 & 66.29 / 32.1 & 57.01 / 20.1\\
 & +\RaTW & 74.90 / 71.7 & 66.48 / 39.2 & 56.93 / 25.6\\
\midrule
TSV-M & Native (uniform) & 74.79 / 62.8 & 62.94 / 31.1 & 53.96 / 22.0\\
 & MetaGPT & 71.22 / 54.6 & 56.21 / 20.4 & 47.15 / 14.1\\
 & NAN & 72.77 / 52.6 & 62.88 / 30.0 & 54.48 / 20.6\\
 & AdaMerging & 75.40 / 63.8 & 63.67 / 27.4 & 54.73 / 20.5\\
 & +\RaTW & 76.54 / 77.3 & 63.94 / 40.3 & 54.30 / 23.9\\
\bottomrule
\end{tabular}
  }
\end{table}

For the three aggregation-based methods in Table~\ref{tab:app-learned}, we also learn task coefficients by minimizing cross-entropy on the same labeled validation data. This supervised reference starts from Native, runs 1{,}000 steps for each of four learning rates, and selects checkpoints on validation. Under the reported search settings, supervised learning exceeds +\RaTW in both metrics for Task Arithmetic on ViT-B/32, whereas +\RaTW achieves higher values in both metrics for TIES and in Mean for TSV-M. The TIES comparison includes trim-threshold selection for +\RaTW; when the threshold is fixed, its tail20 at 50 tasks is below that of supervised learning (Table~\ref{tab:app-learned}, TIES note).

On ViT-L/14 with Task Arithmetic at 50 tasks, a single run of the same supervised procedure reaches 62.59 / 32.1 in Mean / tail20, compared with 59.59 / 24.3 for +\RaTW. At 20 tasks, supervised learning achieves 0.31\,pp higher Mean but 6.3\,pp lower tail20 than +\RaTW.

Supervised coefficient learning requires backpropagation through the merged model, and our implementation keeps all task vectors in GPU memory. It takes 31--68 GPU-minutes per ViT-B/32 setting, compared with an estimated 6--15 minutes for the +\RaTW search, excluding recovery scoring (Table~\ref{tab:app-learned}, search-cost note). These alternative coefficient methods assign one coefficient per task vector and do not apply directly to the six reconstruction-based methods evaluated here, whereas \RaTW also applies to those methods.

\begin{table*}[!tp]
  \centering\footnotesize
  \caption{\textbf{Performance and cost of supervised coefficient learning and training-free weighting rules on ViT-B/32.} The comparison covers three aggregation-based methods with access to the same labeled validation data. Performance is test Mean / tail20 (\%); Cand./steps reports validation-scored candidates or optimization steps, which are not equivalent units of computational cost. Time measures search or training on one RTX 6000 Ada GPU, excluding shared setup and testing; peak allocated GPU memory includes all $T$ task vectors, and both time and memory are listed as $T=50$ / $T=100$. The +\RaTW performance entries repeat Table~\ref{tab:transfer-b32}, with the search scopes and timing qualifications below.
\textbf{Search cost:} Candidate counts cover the $5c$ additional configurations; the $c$ Native configurations are reused. +\RaTW search time is estimated as five times the measured cost of one $\gamma$ search and excludes recovery scoring.
\textbf{TIES:} TIES Native and +\RaTW performance uses the full 71-configuration grid in Table~\ref{tab:app-native-grids}, including trim-threshold selection. Their candidate counts and timings instead cover the 13 output scales at Native's selected threshold. With that threshold fixed, as in the other rows, +\RaTW gives 49.17 / 13.2 at $T=50$ and the tabulated result at $T=100$.
\textbf{Task Arithmetic, $T=50$:} The timing rerun selected a different near-tied candidate, yielding 48.60 / 11.7; the table retains the performance reported in Table~\ref{tab:transfer-b32}.}
  \label{tab:app-learned}\label{tab:app-coef}
  {\setlength{\tabcolsep}{4pt}
\begin{tabular}{@{}llcccccc@{}}
\toprule
 & & \multicolumn{2}{c}{Mean / tail20} & & & \\
\cmidrule(lr){3-4}
Method & Coefficients & $T=50$ & $T=100$ & Cand./steps & Time & Peak GPU mem.\\
\midrule
Task Arithmetic & Native (uniform) & 48.64 / 10.1 & 43.17 / 10.7 & 13 & 1.3 / 2.5 min & 30.7 / 36.2 GB\\
 & +\RaTW & 48.69 / 13.2 & 43.21 / 11.3 & 65 & 6.5 / 12.6 min & 30.7 / 36.2 GB\\
 & Supervised cross-entropy & 49.62 / 13.5 & 44.17 / 12.9 & 4{,}000 steps & 32 / 67 min & 41.1 / 46.4 GB\\
 & MetaGPT & 48.11 / 9.7 & 43.05 / 10.4 & 13 & 1.3 / 2.5 min & 30.7 / 36.2 GB\\
 & NAN & 49.23 / 10.4 & 43.75 / 11.0 & 13 & 1.3 / 2.5 min & 30.7 / 36.2 GB\\
\midrule
TIES & Native (uniform) & 49.21 / 11.0 & 44.01 / 11.5 & 13 & 1.2 / 2.6 min & 30.7 / 36.2 GB\\
 & +\RaTW & 49.51 / 15.8 & 44.05 / 13.4 & 65 & 6.3 / 13.0 min & 30.7 / 36.2 GB\\
 & Supervised cross-entropy & 49.01 / 14.7 & 44.03 / 12.0 & 4{,}000 steps & 31 / 68 min & 41.1 / 46.4 GB\\
 & MetaGPT & 49.03 / 10.4 & 43.94 / 11.3 & 13 & 1.3 / 2.7 min & 30.7 / 36.2 GB\\
 & NAN & 49.27 / 10.7 & 44.14 / 11.3 & 13 & 1.3 / 2.6 min & 30.7 / 36.2 GB\\
\midrule
TSV-M & Native (uniform) & 62.94 / 31.1 & 53.96 / 22.0 & 15 & 1.4 / 2.9 min & 30.7 / 36.2 GB\\
 & +\RaTW & 63.94 / 40.3 & 54.30 / 23.9 & 75 & 7.4 / 14.6 min & 30.7 / 36.2 GB\\
 & Supervised cross-entropy & 62.71 / 40.5 & 52.84 / 23.6 & 4{,}000 steps & 31 / 67 min & 41.1 / 46.4 GB\\
 & MetaGPT & 56.21 / 20.4 & 47.15 / 14.1 & 15 & 1.5 / 2.9 min & 30.7 / 36.2 GB\\
 & NAN & 62.88 / 30.0 & 54.48 / 20.6 & 15 & 1.5 / 2.9 min & 30.7 / 36.2 GB\\
\bottomrule
\end{tabular}
  }
\end{table*}

\subsection{Cost and label budget}
\label{app:cost}
Table~\ref{tab:app-cost} reports computational costs on shared NVIDIA RTX 6000 Ada GPUs (48\,GB). These measurements are GPU-slot hours, not exclusive-runtime comparisons. For the reported Task Arithmetic and TSV-M settings, the full +\RaTW pipeline, including Native search, recovery scoring, and searches over $\gamma$, uses approximately 2.3--2.7 times Native's recorded GPU-slot hours. For settings with complete paired timings, the tabulated three-round \RFpp costs are approximately 5--32\% higher than those of uniform-weight ReForge. This comparison uses the two refinement-cost columns in Table~\ref{tab:app-cost}; initialization and recovery-scoring costs are listed separately. Recovery scoring is reported separately at 0.44--0.76 GPU-slot hours on ViT-L/14; Table~\ref{tab:app-cost} specifies the accounting for shared initialization and scoring.

\begin{table*}[!tp]
\centering\footnotesize
\setlength{\tabcolsep}{6pt}
\caption{\textbf{Recorded computational cost on ViT-L/14.} Entries are GPU-slot hours measured on shared RTX 6000 Ada GPUs and should not be interpreted as exclusive-runtime ratios. \textbf{Upper block:} Native includes the search over the method's grid in Table~\ref{tab:app-native-grids} and final testing, including a parity check that re-merges the selected uniform-weight model with the original implementation and requires validation means to agree within 0.05\,pp. Fixed $\gamma$ pipeline: Native plus recovery scoring and the search at $\gamma=0.5$. Searched $\gamma$ pipeline: the fixed pipeline plus the searches at $\gamma\in\{0.25,1,2,3\}$. \textbf{Lower block:} ReForge and \RFpp costs cover three refinement rounds, with recovery-scoring costs reported separately. Native timings include signal measurement. Initial ISO-CTS construction adds 5.33 / 10.69 hours at 50 / 100 tasks, shared by refinement chains. Recovery-scoring costs include the shared initial measurement job for RegMean and ISO-CTS. \textbf{ISO-CTS timing:} ISO-CTS uses the extended 57-candidate grid; its four $\gamma\neq0.5$ searches ran concurrently on shared GPUs, which inflates its searched-pipeline entry (26.78 hours). The \RFpp entry for 50 \DAs was not fully recorded in the original run; it was measured by rerunning the same three-round chain (same code, settings and BO seed) on one GPU.}
\label{tab:app-cost}
\begin{tabular}{@{}llrrr@{}}
\toprule
Method & $T$ & Native & Fixed $\gamma$ pipeline & Searched $\gamma$ pipeline \\
\midrule
ISO-CTS & 50 & 6.02 & 9.30 & 26.78 \\
Task Arithmetic & 50 & 4.12 & 7.62 & 9.66 \\
Task Arithmetic & 100 & 6.23 & 11.49 & 16.70 \\
TSV-M & 50 & 4.22 & 7.97 & 9.99 \\
\midrule
Setting & & ReForge (3 rounds) & \RFpp (3 rounds) & Recovery scoring only \\
\midrule
50 \DFs & & 19.61 & 25.83 & 0.45 \\
50 \DAs & & 20.10 & 22.73 & 0.44 \\
100 \DFs & & 44.73 & 46.76 & 0.76 \\
100 \DAs & & 43.86 & 48.59 & 0.75 \\
\bottomrule
\end{tabular}
\end{table*}

\label{app:ext-calib}
Table~\ref{tab:app-calib} varies the number of samples used to estimate $\beta$ in two ViT-B/32 settings, fixing $\gamma=0.5$ and ReForge's reconstruction parameters. We draw five random subsets at each budget of 32, 64, and 128 samples per task. At all three budgets, average Mean and tail20 remain close to the results obtained with the full validation subset and exceed those of uniform weighting. Model selection still uses the full validation subset, so this experiment evaluates the sample requirements of recovery-score estimation rather than reducing the end-to-end label budget.

\begin{table}[!tp]
  \centering\footnotesize
  \caption{\textbf{Sample budget for recovery-score estimation.} Entries report test Mean / tail20 (\%) for two ViT-B/32 settings. Only the number of validation images used to estimate $\beta$ varies; $\gamma=0.5$ and ReForge's selected scales and ridge strengths remain fixed. Model selection still uses the full validation subset, so this experiment does not reduce the end-to-end label budget. For each reduced sample size, mean$\pm$std is computed over five random subsets with seeds 0--4, using sample standard deviation ($n-1$ normalization).}
  \label{tab:app-calib}\label{tab:app-lowlabel}
  \setlength{\tabcolsep}{2pt}
  \begin{tabular}{@{}lrrrr@{}}
    \toprule
    & \multicolumn{2}{c}{20 \DAs} & \multicolumn{2}{c}{100 \DFs}\\
    \cmidrule(lr){2-3}\cmidrule(l){4-5}
    Images per task & Mean & tail20 & Mean & tail20\\
    \midrule
    Uniform weights & 80.14 & 74.5 & 61.19 & 29.3\\
    \midrule
    32 & $80.90\pm0.19$ & $80.1\pm0.9$ & $61.44\pm0.03$ & $30.6\pm0.1$\\
    64 & $80.99\pm0.14$ & $80.4\pm0.9$ & $61.44\pm0.02$ & $30.7\pm0.1$\\
    128 & $81.02\pm0.10$ & $80.4\pm0.6$ & $61.45\pm0.02$ & $30.7\pm0.1$\\
    All ($\leq1{,}000$) & 80.99 & 80.3 & 61.46 & 30.7\\
    \bottomrule
  \end{tabular}
\end{table}

\section{Additional ReForge++ Results}
\label{app:reforge}
This section supplements Sec.~\ref{sec:main}.

\subsection{Search-seed stability}
\label{app:ext-seeds}
Table~\ref{tab:app-seeds} compares ReForge and \RFpp at 100 tasks using \DFs and \DAs on both backbones, with three rounds of 200 BO trials per method and only the BO seed varied across runs (seeds 0, 1, and 2). Across all four settings, \RFpp achieves higher three-seed averages in both Mean and tail20 than same-budget ReForge, and both metrics are also higher for every corresponding seed. For example, on ViT-L/14 with \DFs, tail20 is $46.8\pm0.2$ for \RFpp, compared with $39.9\pm0.2$ for ReForge. Across the results in the table, the standard deviation over BO seeds is at most 0.35\,pp for Mean and 0.5\,pp for tail20. Table~\ref{tab:app-seeds-2050} repeats the replication at 20 and 50 tasks. \RFpp again exceeds ReForge in both Mean and tail20 for every seed in all eight settings.

Table~\ref{tab:app-boot-t4} adds test-sampling uncertainty to the matched-budget comparison of Table~\ref{tab:main}: it gives 95\% paired bootstrap intervals for three-round \RFpp minus three-round ReForge (BO seed 0), with the protocol of App.~\ref{app:ext-boot}. All 24 intervals lie above zero. As in App.~\ref{app:ext-boot}, these intervals cover test sampling only; the seed replications above address search randomness.

\begin{table*}[!tp]
\centering\footnotesize
\setlength{\tabcolsep}{3pt}
\caption{\textbf{Paired bootstrap intervals for the matched-budget gains of \RFpp} (rows ``\RFpp, 3'' minus ``ReForge, 3'' of Table~\ref{tab:main}, BO seed 0; pp; 95\% percentile intervals in brackets, 10{,}000 resamples of the test images within each task). Point estimates are computed from unrounded results.}
\label{tab:app-boot-t4}
\begin{tabular}{@{}lcccccc@{}}
\toprule
 & \multicolumn{2}{c}{$T=20$} & \multicolumn{2}{c}{$T=50$} & \multicolumn{2}{c}{$T=100$}\\
\cmidrule(lr){2-3}\cmidrule(lr){4-5}\cmidrule(l){6-7}
Backbone, statistics & $\Delta$Mean & $\Delta$tail20 & $\Delta$Mean & $\Delta$tail20 & $\Delta$Mean & $\Delta$tail20\\
\midrule
ViT-L/14, \DFs & +0.74 {\scriptsize[+0.65, +0.84]} & +3.4 {\scriptsize[+3.1, +3.7]} & +0.94 {\scriptsize[+0.82, +1.05]} & +6.4 {\scriptsize[+5.5, +6.8]} & +1.34 {\scriptsize[+1.23, +1.45]} & +6.8 {\scriptsize[+6.3, +7.4]}\\
ViT-L/14, \DAs & +0.91 {\scriptsize[+0.82, +1.00]} & +4.5 {\scriptsize[+4.2, +4.7]} & +1.07 {\scriptsize[+0.96, +1.19]} & +5.6 {\scriptsize[+5.1, +6.1]} & +1.08 {\scriptsize[+0.95, +1.20]} & +5.6 {\scriptsize[+5.1, +6.1]}\\
ViT-B/32, \DFs & +1.92 {\scriptsize[+1.79, +2.05]} & +10.0 {\scriptsize[+9.5, +10.5]} & +1.37 {\scriptsize[+1.19, +1.54]} & +7.0 {\scriptsize[+6.2, +7.9]} & +0.84 {\scriptsize[+0.71, +0.96]} & +3.8 {\scriptsize[+3.3, +4.1]}\\
ViT-B/32, \DAs & +1.14 {\scriptsize[+1.03, +1.25]} & +6.2 {\scriptsize[+5.8, +6.6]} & +1.06 {\scriptsize[+0.93, +1.19]} & +7.4 {\scriptsize[+6.9, +7.9]} & +0.56 {\scriptsize[+0.45, +0.66]} & +2.9 {\scriptsize[+2.7, +3.2]}\\
\bottomrule
\end{tabular}
\end{table*}

\begin{table}[!tp]
\centering\footnotesize
\setlength{\tabcolsep}{3pt}
\caption{\textbf{BO-seed replication at 100 tasks.} Both backbones use three rounds of 200 BO trials each. Entries report the final round-3 model's test Mean and tail20 (\%); within each method and setting, only the BO seed varies. Mean$\pm$std summarizes seeds 0, 1 and 2 using sample standard deviation ($n-1$ normalization).}
\label{tab:app-seeds}
\resizebox{\linewidth}{!}{%
\begin{tabular}{@{}llrrrr@{}}
\toprule
& & \multicolumn{2}{c}{ReForge} & \multicolumn{2}{c}{\RFpp} \\
\cmidrule(lr){3-4}\cmidrule(l){5-6}
Setting & Seed & Mean & tail20 & Mean & tail20 \\
\midrule
\multicolumn{6}{@{}l}{\textit{ViT-B/32}}\\
100 \DFs & 0 & 61.57 & 30.2 & 62.41 & 34.0 \\
 & 1 & 61.58 & 30.1 & 62.70 & 34.7 \\
 & 2 & 61.62 & 30.3 & 62.57 & 34.1 \\
 & mean$\pm$std & $61.59\pm0.03$ & $30.2\pm0.1$ & $62.56\pm0.15$ & $34.3\pm0.4$ \\
\midrule
100 \DAs & 0 & 63.75 & 33.5 & 64.31 & 36.4 \\
 & 1 & 63.78 & 33.3 & 64.44 & 36.1 \\
 & 2 & 63.72 & 33.1 & 63.78 & 35.6 \\
 & mean$\pm$std & $63.75\pm0.03$ & $33.3\pm0.2$ & $64.17\pm0.35$ & $36.0\pm0.4$ \\
\midrule
\multicolumn{6}{@{}l}{\textit{ViT-L/14}}\\
100 \DFs & 0 & 73.20 & 40.1 & 74.54 & 46.9 \\
 & 1 & 73.13 & 39.9 & 74.34 & 46.6 \\
 & 2 & 73.13 & 39.7 & 74.21 & 46.9 \\
 & mean$\pm$std & $73.15\pm0.04$ & $39.9\pm0.2$ & $74.36\pm0.17$ & $46.8\pm0.2$ \\
\midrule
100 \DAs & 0 & 75.63 & 46.4 & 76.70 & 51.9 \\
 & 1 & 76.03 & 47.4 & 76.75 & 51.6 \\
 & 2 & 75.71 & 46.8 & 76.72 & 51.7 \\
 & mean$\pm$std & $75.79\pm0.21$ & $46.9\pm0.5$ & $76.72\pm0.03$ & $51.7\pm0.1$ \\
\bottomrule
\end{tabular}}
\end{table}

\begin{table}[!tp]
\centering\footnotesize
\setlength{\tabcolsep}{3pt}
\caption{\textbf{BO-seed replication at 20 and 50 tasks.} Same protocol as Table~\ref{tab:app-seeds}; entries are mean$\pm$std of round-3 test Mean / tail20 (\%) over BO seeds 0, 1 and 2.}
\label{tab:app-seeds-2050}
\resizebox{\linewidth}{!}{%
\begin{tabular}{@{}llcc@{}}
\toprule
Backbone & Setting & ReForge & \RFpp \\
\midrule
ViT-B/32 & 20 \DFs & $80.53\pm0.07$ / $76.1\pm1.3$ & $82.31\pm0.08$ / $86.3\pm0.3$ \\
 & 20 \DAs & $81.97\pm0.10$ / $80.4\pm0.0$ & $83.05\pm0.02$ / $87.0\pm0.5$ \\
 & 50 \DFs & $70.70\pm0.13$ / $48.9\pm0.7$ & $71.93\pm0.26$ / $56.2\pm0.3$ \\
 & 50 \DAs & $73.41\pm0.18$ / $51.9\pm0.4$ & $74.46\pm0.20$ / $59.3\pm0.8$ \\
\midrule
ViT-L/14 & 20 \DFs & $90.21\pm0.05$ / $90.1\pm0.5$ & $91.04\pm0.06$ / $94.2\pm0.5$ \\
 & 20 \DAs & $90.21\pm0.06$ / $89.5\pm0.4$ & $91.14\pm0.04$ / $94.4\pm0.3$ \\
 & 50 \DFs & $84.43\pm0.09$ / $75.0\pm0.4$ & $85.30\pm0.03$ / $80.6\pm0.3$ \\
 & 50 \DAs & $85.24\pm0.14$ / $75.3\pm0.2$ & $86.06\pm0.10$ / $80.6\pm0.1$ \\
\bottomrule
\end{tabular}}
\end{table}

\subsection{Recomputing recovery scores}
\label{app:fixed-beta}
Table~\ref{tab:app-fixed-beta} tests recovery-score recomputation at 100 tasks, using three rounds of 200 BO trials each. The variants share round 1; the fixed-score variant reuses the scores measured at the initial model in rounds 2 and 3, whereas \RFpp recomputes them at each round's reference. Both variants update the anchor and search $\gamma$ in every round; the ViT-B/32 comparisons use seed 0, and the ViT-L/14 comparisons use seed 1 with \DAs and \DFs. Fixed recovery scores therefore do not imply fixed task weights.

On ViT-B/32 with \DAs, fixed and recomputed scores give similar Mean / tail20 results: 64.29 / 36.2 and 64.31 / 36.4, respectively. With \DFs, fixed scores yield higher values (62.70 / 35.1 versus 62.41 / 34.0). On ViT-L/14 with \DAs, fixed scores also yield higher values (76.91 / 52.3 versus 76.75 / 51.6), whereas with \DFs recomputed scores yield higher values (74.34 / 46.6 versus 74.21 / 45.8). These runs do not establish a consistent advantage from recomputation, although both variants outperform same-budget uniform-weight ReForge in both metrics in all four comparisons.

Table~\ref{tab:app-fixed-beta}(b) reports $\gamma$ for the model selected in each round. Both recomputed ViT-B/32 chains and both recomputed ViT-L/14 \DFs chains (seeds 0 and 1) select $\gamma=0$ in rounds 2 and 3, using uniform task weights while continuing to refine the inherited anchor. This pattern is not universal: the seed-0 ViT-L/14 \DAs chain selects positive $\gamma$ in all three rounds, and three of the four fixed-score chains select a positive value in at least one later round. Later-round refinement can therefore proceed with uniform task weights; $\gamma=0$ describes the selected model, not necessarily all candidates evaluated in that round.

\begin{table*}[!tp]
\centering\footnotesize
\caption{\textbf{Fixed versus recomputed recovery scores.} All settings use $T=100$ and three rounds of 200 BO trials each. The fixed-score and recomputed-score chains share round 1 and differ in whether recovery scores are recomputed in subsequent rounds. Both variants search $\gamma$ in every round, so fixed scores do not imply fixed task weights. \textbf{(a)} Test Mean / tail20 (\%) of the round-3 model: ViT-B/32 uses BO seed 0 with $\gamma\in[0,3]$, and ViT-L/14 uses BO seed 1 with $\gamma\in[0,2]$. \textbf{(b)} Recovery exponent $\gamma$ of the model selected by BO in each round, using BO seed 0 unless stated otherwise. Entries of 0.00 are exactly zero and indicate uniform task weights for the selected model, not necessarily for all candidates evaluated in that round.}
\label{tab:app-fixed-beta}\label{tab:app-gamma}
\begin{minipage}[t]{0.56\textwidth}\centering
(a) Round-3 results\par\smallskip
\resizebox{\linewidth}{!}{%
\begin{tabular}{@{}lcccc@{}}
\toprule
 & B/32, \DAs & B/32, \DFs & L/14, \DAs (seed 1) & L/14, \DFs (seed 1)\\
\midrule
Uniform (ReForge) & 63.75 / 33.5 & 61.57 / 30.2 & 76.03 / 47.4 & 73.13 / 39.9\\
$\beta$ fixed after round 1 & 64.29 / 36.2 & 62.70 / 35.1 & 76.91 / 52.3 & 74.21 / 45.8\\
$\beta$ recomputed (\RFpp) & 64.31 / 36.4 & 62.41 / 34.0 & 76.75 / 51.6 & 74.34 / 46.6\\
\bottomrule
\end{tabular}}
\end{minipage}\hfill
\begin{minipage}[t]{0.42\textwidth}\centering
(b) Selected $\gamma$\par\smallskip
\resizebox{\linewidth}{!}{%
\begin{tabular}{@{}lccc@{}}
\toprule
Chain & Round 1 & Round 2 & Round 3\\
\midrule
ViT-B/32, \DAs, \RFpp & 1.87 & 0.00 & 0.00\\
ViT-B/32, \DFs, \RFpp & 1.54 & 0.00 & 0.00\\
ViT-B/32, \DAs, fixed $\beta$ & 1.87 & 0.47 & 0.00\\
ViT-B/32, \DFs, fixed $\beta$ & 1.54 & 1.41 & 1.06\\
ViT-L/14, \DFs, \RFpp (seed 0) & 1.10 & 0.00 & 0.00\\
ViT-L/14, \DAs, \RFpp (seed 0) & 1.57 & 0.66 & 0.25\\
ViT-L/14, \DAs, \RFpp (seed 1) & 1.25 & 0.33 & 0.00\\
ViT-L/14, \DAs, fixed $\beta$ (seed 1) & 1.25 & 0.35 & 1.16\\
ViT-L/14, \DFs, \RFpp (seed 1) & 1.24 & 0.00 & 0.00\\
ViT-L/14, \DFs, fixed $\beta$ (seed 1) & 1.24 & 0.00 & 0.00\\
\bottomrule
\end{tabular}}
\end{minipage}
\end{table*}

\input{sec/app_vlm}

%% file: sec/X_appendix_B.tex
\section{Method Details: \RaTW Instantiations}
\label{app:derive}

\subsection{Aggregation: processed task contributions}
\label{app:paradigms}
This section describes how \RaTW weights task contributions in five aggregation-based merging methods, using the common form introduced in Sec.~\ref{sec:rw}:
\begin{equation}
 \tau_{\rm merge}(q)=\sum_tq_t\widehat\tau_t(q),\qquad
 \theta_{\rm merge}=\theta_0+\eta\tau_{\rm merge}(q).
 \label{eq:app-aggregation-map}
\end{equation}
Here, $\widehat\tau_t(q)$ is the task contribution after method-specific processing, $q_t$ is its task weight, and $\eta$ is the native output scale. The processed contributions may themselves depend on $q=(q_1,\ldots,q_T)^\top$ through normalization, so the overall merge need not be linear in $q$. The output scale $\eta$ corresponds to $s$ in Task Arithmetic, TIES and ISO-CTS, and to $\alpha$ in TSV-M (Table~\ref{tab:app-native-grids}). Fisher merging uses $\eta=1$.

\paragraph{Task Arithmetic.}
The task contribution is the original task vector, $\widehat\tau_t=\tau_t$, which \RaTW weights directly in the sum. The native output scale $\eta=s$ remains separate, with no additional division by $T$.

\paragraph{Fisher merging.}
\RaTW weights each task's contribution to both the numerator and denominator of the Fisher merge. Let $F_{t,i}$ denote the diagonal Fisher estimate at parameter $i$; the native procedure adds epsilon to obtain $\widetilde F_{t,i}=F_{t,i}+\epsilon$. When optional task-level Fisher normalization is enabled, its coefficients are computed as
\begin{equation}
 n_t=\left(\sum_iF_{t,i}^{2}\right)^{1/2},\qquad
 a_t=\frac{(n_t+\epsilon)^{-1}}{\sum_j(n_j+\epsilon)^{-1}};
 \label{eq:app-fisher-norm}
\end{equation}
otherwise, $a_t=1$. Combining these quantities with the native coefficient $b_t=1/T$ gives $h_{t,i}=b_ta_t\widetilde F_{t,i}$, and the weighted merge is
\begin{align}
 Z_i(q)&=\sum_jq_jh_{j,i},\nonumber\\
 \theta_{{\rm merge},i}&=\frac{\sum_tq_th_{t,i}\theta_{t,i}}{Z_i(q)},\qquad
 \widehat\tau_{t,i}(q)=\frac{h_{t,i}\tau_{t,i}}{Z_i(q)}.
 \label{eq:app-fisher-map}
\end{align}
The optional task-level normalization is computed before applying the \RaTW weights.

\paragraph{TIES.}
Native whole-vector trimming and sign election operate on unweighted task vectors, so \RaTW does not change the elected signs. Let $z_{t,i}$ denote a retained sign-consistent entry and $m_{t,i}=\mathbf1[z_{t,i}\ne0]$ its support indicator. The hyperparameter search includes disjoint sum, which weights the retained entries directly:
\begin{equation}
 \widehat\tau_{t,i}=z_{t,i},\qquad \tau_{{\rm merge},i}=\sum_tq_tz_{t,i},
 \label{eq:app-ties-sum}
\end{equation}
It also includes disjoint mean, which normalizes by the weighted support mass:
\begin{equation}
 M_i(q)=\sum_jq_jm_{j,i},\qquad
 \widehat\tau_{t,i}(q)=\begin{cases}z_{t,i}/M_i(q),&M_i(q)>0,\\0,&M_i(q)=0.\end{cases}
 \label{eq:app-ties-mean}
\end{equation}
This rule returns zero when the support mass is zero and uses the actual denominator whenever it is positive, including values below one.

\paragraph{TSV-M.}
\RaTW weights the task contributions constructed by native TSV-M. For a matrix block $V_t\in\mathbb R^{m\times n}$, set $r=\min(m,n)$ and $d=\lfloor r/T\rfloor$. The native procedure retains $d$ components from each task's unweighted thin SVD and concatenates their factors, with the right singular vectors stored as rows of $H_t$:
\begin{equation}
 \begin{aligned}
 V_t&=U_t\operatorname{diag}(s_t)H_t,\\
 U&=[U_{1,:d},\ldots,U_{T,:d}],\quad
 H=[H_{1,:d}^{\top},\ldots,H_{T,:d}^{\top}]^{\top}.
 \end{aligned}
 \label{eq:app-tsv-factors}
\end{equation}
Arrays and singular values are zero-padded to the native width $r$ when $Td<r$. For a thin SVD $B=L_B\Sigma_BR_B^\top$, define $\mathcal P(B)=L_BR_B^\top$, giving the native polar factors $\bar U=\mathcal P(U)$ and $\bar H=\mathcal P(H)$.

For the index block $I_t$ containing task $t$'s retained components, its processed contribution and the weighted merge are
\begin{equation}
 \widehat V_t=\bar U_{:,I_t}\operatorname{diag}(s_{t,:d})\bar H_{I_t,:},\qquad
 V_{\rm merge}=\sum_tq_t\widehat V_t.
 \label{eq:app-tsv-map}
\end{equation}
Thus, $q_t$ scales the retained singular-value block without recomputing the bases or polar factors. Non-matrix parameters use the weighted mean $\sum_tq_t\tau_t/T$, corresponding to $\widehat\tau_t=\tau_t/T$.

\paragraph{ISO-CTS.}
\RaTW retains the native common block once and weights only the task-specific blocks. With the matrix dimensions and rank defined above, the native common-space fraction $\rho$ determines the rank allocation:
\begin{equation}
 d=\operatorname{round}\!\left(\frac{r-\lfloor r\rho\rfloor}{T}\right),\qquad c=r-Td.
 \label{eq:app-iso-ranks}
\end{equation}
The leading $c$ singular components of $\sum_tV_t$ define the common factors $U_c,s_c,H_c$. Each task contributes the leading $d$ components of its residual after projection out of the common left subspace:
\begin{equation}
 (I-U_cU_c^\top)V_t=U_t^\perp\operatorname{diag}(s_t^\perp)H_t^\perp.
 \label{eq:app-iso-residual}
\end{equation}
The task-specific factors are concatenated first, followed by the common factors, and the concatenated factors are polar-orthogonalized to obtain $\bar U,\bar H$. Native isotropization uses the shared singular value
\begin{equation}
 \bar s=\frac{\sum_t\sum_{j=1}^{d}s_{t,j}^\perp+\sum_{j=1}^{c}s_{c,j}}{r}
 \label{eq:app-iso-isotropy}
\end{equation}
to construct the common and task-specific blocks:
\begin{equation}
 C=\bar s\bar U_{:,I_c}\bar H_{I_c,:},\qquad S_t=\bar s\bar U_{:,I_t}\bar H_{I_t,:},
 \label{eq:app-iso-parts}
\end{equation}
where $I_c$ and $I_t$ index their respective components.

The weighted update is
\begin{equation}
 V_{\rm merge}=C+\sum_tq_tS_t=\sum_tq_t\underbrace{(S_t+C/T)}_{\widehat V_t}.
 \label{eq:app-iso-map}
\end{equation}
The second equality follows from $\sum_tq_t=T$ and expresses the update in the common form of Eq.~\eqref{eq:app-aggregation-map}. Weighting does not recompute the bases or $\bar s$, nor is isotropization repeated afterward. Non-matrix parameters use the TSV-M weighted mean, dividing the weighted sum by the number of tasks. The native output scale multiplies the complete update.

\subsection{Reconstruction: task terms and native solvers}
\label{app:methods}
This section describes how \RaTW weights task-fitting contributions in reconstruction-based methods while retaining their native solvers. For one unit, let $W_0,W_t$ denote the pre-trained and expert weights, with expert update $V_t=W_t-W_0$. The merged weight in full-weight coordinates is $U=W_0+\tau$. Each method solves in its native coordinates and applies output scaling afterward.

\paragraph{RegMean.}
\RaTW multiplies each task's Gram matrix by $q_t$ while retaining the native RegMean solve. The unshrunk Gram matrix $G_t^{\rm raw}=X_t^\top X_t/N_t^{\rm in}$ averages over expert-input rows. Native off-diagonal shrinkage then gives
\[
\widetilde G_t=\alpha G_t^{\rm raw}+(1-\alpha)\operatorname{Diag}(G_t^{\rm raw}),
\]
where $\alpha$ is the shrinkage factor and $\widetilde G_t$ corresponds to $G_t$ in Sec.~\ref{sec:rw}. RegMean uses Eq.~\eqref{eq:quadratic_fit} with $\Omega=0$; in the shared solve of Eq.~\eqref{eq:app-closed}, $Y_t=W_t$, $G_t=\widetilde G_t$ and $\lambda=0$, and the returned update is $\tau=U-W_0$. Task weighting commutes with shrinkage, and non-regressed parameters retain native handling.

\paragraph{RegMean++.}
RegMean++ constructs its statistics from inputs that have passed through the current merged prefix. At block $\ell$, task inputs first pass through that prefix and then enter expert block $\ell$, where submodule $j$ collects $X_{t,\ell,j}^{++}$ and forms the native averaged, shrunk Gram matrix $\widetilde G_t^{++}$. Suppressing unit indices, the fitting term is
\begin{equation}
 \mathcal D_t(\tau)=\operatorname{tr}[(\tau-V_t)\widetilde G_t^{++}(\tau-V_t)^\top],\quad \Omega=0,
 \label{eq:app-rmpp}
\end{equation}
with full-weight target $W_{t,\ell,j}$ and corresponding output target $X_{t,\ell,j}^{++}W_{t,\ell,j}^\top$. \RaTW multiplies each Gram matrix by $q_t$ before the native closed-form solve. The newly merged block supplies the next block's inputs, so statistics are fixed within each solve but can differ between Native and +\RaTW. Non-regressed parameters retain native averaging.

\paragraph{LOT.}
Expert input activations from the selected exemplars define the fitting term
\begin{equation}
 \mathcal D_t(\tau)=\|X_t(\tau-V_t)^\top\|_F^2,\qquad \Omega=0.
 \label{eq:app-lot}
\end{equation}
With $G_t=X_t^\top X_t$, \RaTW weights each task's contribution to both the target--Gram products and the Gram matrices. The resulting solution is
\[
\tau^\star=(\sum_tq_tV_tG_t)(\sum_tq_tG_t)^\dagger.
\]
Exemplar selection remains unchanged.

For normalization-scale vectors $v_t$, native normalized features $z_{t,a}$ give $g_{t,i}=\sum_a z_{t,a,i}^2$. Their fitting objective is
\[
\mathcal D_t(u)=\sum_i g_{t,i}(u_i-v_{t,i})^2.
\]
The weighted solution is $u_i^\star=\sum_tq_tg_{t,i}v_{t,i}/\sum_tq_tg_{t,i}$, with the native zero-denominator rule. Other parameter handling and deployment scaling $W_0+s\tau^\star$ remain native. The scaled update need not minimize the unscaled objective.

\paragraph{WUDI.}
For nonzero task-vector blocks, the native normalized fitting term is
\begin{equation}
 \mathcal D_t(\tau)=\frac{\|(\tau-V_t)V_t^\top\|_F^2}{\|V_t\|_F^2},\qquad \Omega=0.
 \label{eq:app-wudi}
\end{equation}
This corresponds to Eq.~\eqref{eq:app-dt} with $Y_t=V_t$ and $G_t=V_t^\top V_t/\|V_t\|_F^2$. \RaTW multiplies the entire normalized task term by $q_t$. Optimization retains the native initialization $\sum_tV_t$ and Adam with learning rate $10^{-5}$. The iteration count is selected on the validation set from the native grid spanning 100--12,800 iterations. This finite-step optimization need not reach the exact minimizer; other parameters retain native treatment.

\paragraph{ACE-Merging.}
The full native ACE-Merging procedure includes corrections beyond a scalar quadratic objective. For row-centered blocks $\bar V_t=V_t-\mathbf1\mu_t^\top$, define $G_t=\bar V_t^\top\bar V_t$. The native heterogeneity gate is determined per block from the unweighted task vectors: it is on when the variance of $\log\|V_t\|_F^2$ across tasks, divided by its squared mean, exceeds the native threshold. We set $c_t=1$ with the gate off and $c_t=\operatorname{tr}(G_t)+10^{-12}$ with it on, giving $S_t=G_t/c_t$ and $H_t=S_t+(\epsilon/c_t)I$. These quantities define the covariance-fit core:
\[
\mathcal D_t^{\rm core}(\tau)=\operatorname{tr}[(\tau-\bar V_t)H_t(\tau-\bar V_t)^\top].
\]

The implementation obtains $\tau_{\rm core}$ by solving the corrected system
\begin{equation}
 \tau_{\rm core}\left(\sum_tq_tH_t+C_q\right)=\sum_tq_t\bar V_tH_t.
 \label{eq:app-ace}
\end{equation}
Each row of $C_q$ is the vector of column means of $\sum_tq_tS_t$, additionally divided by the native mean of $\|V_t\|_F^2$ when the gate is on. The solver uses the native inverse with pseudo-inverse fallback. Because $C_q$ is generally nonsymmetric, its correction cannot be represented by a scalar quadratic penalty $\Omega$.

With the gate on, the procedure computes $\bar H_q=\sum_tq_tH_t/\sum_tq_t$ and $\Delta_q=\sum_tq_tV_t(S_t-\bar H_q)$. The SVD of $\tau_{\rm core}+\Delta_q$ supplies the native top-$k$ left/right singular-vector product. This product is scaled by the mean retained singular value and added to $\tau_{\rm core}$. \RaTW weights the task-dependent sums above while preserving the native gate, rank fraction, regularization and output scale. Non-matrix parameters use $\sum_tq_tV_t/\sum_tq_t$.

\paragraph{ReForge and \RFpp.}
Both methods use the reconstruction objective in Sec.~\ref{sec:rw}, with expert update $V_{t,m}$ as the target and anchor block $A_m$, taken from the anchor's task vector $\tau_A$, as the prior center. \RaTW weights the expert-fitting terms without weighting the anchor penalty. For CLIP, each encoder layer has four solved units---stacked Q/K/V projections, attention output and two MLP weights---giving 48 units for ViT-B/32 and 96 for ViT-L/14. \DAs constructs $X_{t,m}$ from the same 128 training images per task in every round. \DFs uses task-vector statistics and no images.

Encoder layers are divided into three segments, each with a shared deployment scale, while ridge strengths vary by segment and unit type (Table~\ref{tab:app-space}). Solved updates replace the corresponding anchor task-vector blocks. Other encoder parameters retain the anchor update before segment scaling. Parameters outside the encoder retain unscaled anchor values. Consequently, encoder updates that are not re-solved can accumulate scaling across rounds.

\subsection{Shared solve and numerical handling}
\label{app:quad}
The quadratic fitting terms in Sec.~\ref{app:methods} share the following formulation. For RegMean and RegMean++, $U$ and $Y_t$ denote full unit weights: $U=W_0+\tau$ and $Y_t=W_t$, so $U-Y_t=\tau-V_t$, as in Eq.~\eqref{eq:app-rmpp}. For LOT, WUDI and ReForge, these variables denote task-vector blocks. With symmetric positive-semidefinite $G_t$, define the task-fitting term as
\begin{equation}
 \mathcal D_t(U)=\operatorname{tr}[(U-Y_t)G_t(U-Y_t)^\top].
 \label{eq:app-dt}
\end{equation}
The weighted objective is
\begin{equation}
 J(U)=\sum_tq_t\mathcal D_t(U)+\lambda\|U-A\|_F^2,
 \label{eq:app-j}
\end{equation}
where $\lambda\ge0$ and the local anchor $A$ uses the same coordinates as $U$.

This objective yields the normal equation
\begin{equation}
 U^\star\left(\sum_tq_tG_t+\lambda I\right)=\sum_tq_tY_tG_t+\lambda A.
 \label{eq:app-closed}
\end{equation}
For $\lambda>0$, the solution is unique; at $\lambda=0$, native inverse or pseudo-inverse conventions apply. WUDI uses finite-step optimization of this quadratic, while ACE uses the corrected system described in Sec.~\ref{app:methods}. With fixed task statistics and positive weights, the weighted sum of Gram matrices has the same null space as the unweighted sum. Large weight ratios can nevertheless worsen conditioning. The mean-one normalization in Eq.~\eqref{eq:q} preserves the total task weight of the native uniform setting, avoiding a global rescaling of the data term solely due to a change in total weight.

\paragraph{Numerical handling.}
\label{app:numerics}
The count floor in Eq.~\eqref{eq:beta} keeps task weights positive. When $k_t=0$, it gives $\beta_t=1/|\mathcal C_t|$, so the score still depends on the number of samples used for recovery estimation. For two tasks with zero counts and 70 versus 1{,}000 CLIP validation images (Table~\ref{tab:app-l14-tasks}), the floored scores differ by $14.3\times$, giving a weight ratio of $3.8\times$ under common normalization at $\gamma=0.5$. The VLM experiments apply the same floor to a continuous count, as detailed in Sec.~\ref{app:vlm-setup}.

At the initial model $\theta^{(0)}$ of ViT-L/14 \RFpp, zero counts occur for 0 of 20, 0 of 50 and 3 of 100 tasks. The three tasks are fruit\_ripeness, usps and wonders\_of\_world, with 400, 727 and 270 validation images, respectively. Non-finite values stop the run. The denominator $\|\tau_t\|_F^2$ of $G_t^{\mathrm{DF}}$ is floored at $10^{-12}$.

\paragraph{Equal-weight reproduction.}
\label{app:equal}
At $q\equiv1$, the formulas recover their native rules. Synthetic checks agree within floating-point tolerance. RegMean and RegMean++ also satisfy the weighted normal equations. Re-merging the selected Native models for ACE-Merging, Task Arithmetic, TIES, TSV-M and ISO-CTS reproduces them within $10^{-4}$ in parameters and 0.05 pp in validation mean.

The ReForge checks re-solve one round in four ViT-B/32 settings, holding the selected scales and ridge strengths fixed. ReForge's different accumulation order yields parameter differences of at most $7.2\times10^{-5}$. The validation-accuracy offsets are $-0.16$ pp for \DAs at 20 tasks and $-0.02$ pp for \DFs at 100 tasks.

%% file: sec/tab_tasks.tex
\begin{table*}[!tp]
  \centering\scriptsize
  \setlength{\tabcolsep}{2.6pt}
  \caption{\textbf{CLIPMerge-100 tasks, provenance and expert accuracies.} Tasks 1--50 are listed here; tasks 51--100 follow in the continuation. \textbf{Columns:} Set denotes the smallest task set containing a task. L/14 and B/32 report expert test accuracies (\%), used to normalize retention. Sub. is the size of the fixed validation subset, capped at 1{,}000 images. \textbf{Sources:} Source identifies the Hugging Face dataset unless stated otherwise; dataset revisions are pinned in the release. Prefix abbreviations are cb: = \texttt{clip-benchmark/wds\_}, cbv: = \texttt{clip-benchmark/wds\_vtab-}, tg/ = \texttt{tanganke/}, and dpdl/ = \texttt{dpdl-benchmark/}. \textbf{Splits:} Test codes are P, published test split; V, source split named ``validation'' used for testing; H, an 80/20 split of a single published split by a fixed hash of (label, key); and S, an 80/20 split by image SHA-1, applied after pooling the unstratified published splits for mit\_indoor67 and pokemon, or to the labeled training split for human\_action. Val codes are R, a stratified 10\% of the training split with seed 0; P, a published validation split; and G, a held-out toy instance. \textbf{Duplicates:} Dups reports exact duplicates removed across splits, in the order train$\cap$test / train$\cap$val / val$\cap$test, identified by SHA-1 of stored image bytes or pixel arrays for MedMNIST. ND is the percentage of sampled training images---at most 20{,}000 checked---with a test image at dHash Hamming distance $\le2$. These images are retained, and matches have not been inspected by hand. High match rates on small synthetic or grayscale images, such as dSprites (100\%) and MNIST (51\%), mean that ND indicates similarity rather than confirmed duplication. $^{\dagger}$A published validation split replaced an earlier random split. Dups refers to the original split construction; exact matches to training or test images were subsequently removed from the replacement validation split: 30 for eeg\_seizure\_spectrogram, 1 for person\_presence, and 0 otherwise. $^{\ddagger}$Validation uses toy instance 9; ND was measured before training was restricted to instances 4, 6, 7 and 8.}
  \label{tab:app-provenance}\label{tab:app-l14-tasks}\label{tab:app-tasks}
  \begin{tabular}{@{}lclccrrrrrrrr@{}}
    \toprule
    & & & \multicolumn{2}{c}{Split rule} & \multicolumn{3}{c}{Split size} & & & \multicolumn{2}{c}{Expert acc.} & \\
    \cmidrule(lr){4-5}\cmidrule(lr){6-8}\cmidrule(lr){11-12}
    Task & Set & Source & Test & Val & Train & Val & Test & Dups & ND & L/14 & B/32 & Sub. \\
    \midrule
    sun397 & 20 & tg/sun397 & P & R & 17{,}865 & 1{,}985 & 19{,}850 & 0/0/0 & 0.3 & 82.22 & 73.27 & 1000 \\
    stanford-cars & 20 & tg/stanford\_cars & P & R & 7{,}337 & 784 & 8{,}041 & 15/4/4 & 0.2 & 92.07 & 73.29 & 784 \\
    resisc45 & 20 & tg/resisc45 & P & R & 17{,}008 & 1{,}890 & 6{,}300 & 1/1/0 & 0.0 & 97.08 & 94.78 & 1000 \\
    eurosat & 20 & tg/eurosat & P & R & 19{,}440 & 2{,}160 & 2{,}700 & 0/0/0 & 2.7 & 99.44 & 99.11 & 1000 \\
    svhn & 20 & ufldl-stanford/svhn & P & R & 65{,}931 & 7{,}326 & 26{,}032 & 0/0/0 & 3.4 & 97.87 & 97.05 & 1000 \\
    gtsrb & 20 & tg/gtsrb & P & R & 23{,}976 & 2{,}664 & 12{,}630 & 0/0/0 & 4.0 & 99.15 & 98.74 & 1000 \\
    mnist & 20 & ylecun/mnist & P & R & 54{,}001 & 5{,}999 & 10{,}000 & 0/0/0 & 51.4 & 99.72 & 99.66 & 1000 \\
    dtd & 20 & tg/dtd & P & R & 3{,}375 & 376 & 1{,}880 & 7/2/0 & 4.4 & 84.57 & 76.97 & 376 \\
    oxford\_flowers102 & 20 & dpdl/oxford\_flowers102 & P & R & 917 & 102 & 6{,}149 & 1/0/0 & 0.1 & 98.18 & 87.66 & 102 \\
    pcam & 20 & 1aurent/PatchCamelyon & P & P$^\dagger$ & 215{,}954 & 32{,}768 & 32{,}768 & 32/19{,}944/3 & 0.2 & 90.44 & 87.53 & 1000 \\
    fer2013 & 20 & cb:fer2013 & P & R & 24{,}999 & 2{,}800 & 7{,}178 & 631/207/72 & 2.0 & 75.40 & 69.98 & 1000 \\
    oxford-iiit-pet & 20 & timm/oxford-iiit-pet & P & R & 3{,}311 & 369 & 3{,}669 & 0/0/0 & 0.7 & 96.59 & 92.15 & 369 \\
    stl10 & 20 & tg/stl10 & P & R & 4{,}500 & 500 & 8{,}000 & 0/0/0 & 0.3 & 99.56 & 98.22 & 500 \\
    cifar100 & 20 & uoft-cs/cifar100 & P & R & 44{,}988 & 4{,}999 & 10{,}000 & 9/3/1 & 0.8 & 93.03 & 88.35 & 1000 \\
    cifar10 & 20 & uoft-cs/cifar10 & P & R & 45{,}000 & 5{,}000 & 10{,}000 & 0/0/0 & 0.1 & 99.02 & 97.50 & 1000 \\
    food101 & 20 & ethz/food101 & V & R & 68{,}173 & 7{,}575 & 25{,}250 & 2/0/0 & 0.1 & 95.31 & 88.38 & 1000 \\
    fashion\_mnist & 20 & zalando-datasets/fashion\_mnist & P & R & 54{,}000 & 6{,}000 & 10{,}000 & 0/0/0 & 36.6 & 95.71 & 94.87 & 1000 \\
    emnist\_letters & 20 & tg/emnist\_letters & P & R & 112{,}317 & 12{,}480 & 20{,}800 & 0/3/0 & 24.0 & 95.80 & 95.47 & 1000 \\
    kmnist & 20 & tg/kmnist & P & R & 54{,}000 & 6{,}000 & 10{,}000 & 0/0/0 & 11.0 & 98.75 & 98.33 & 1000 \\
    rendered-sst2 & 20 & cb:renderedsst2 & P & R & 6{,}224 & 690 & 1{,}821 & 1/3/2 & 90.8 & 84.13 & 71.77 & 690 \\
    caltech101 & 50 & cbv:caltech101 & P & R & 2{,}463 & 290 & 6{,}085 & 0/0/0 & 1.3 & 96.71 & 95.02 & 290 \\
    clevr\_distance & 50 & cbv:clevr\_closest\_object\_distance & P & R & 56{,}700 & 6{,}300 & 15{,}000 & 0/0/0 & 0.5 & 88.80 & 86.59 & 1000 \\
    dsprites\_orientation & 50 & cbv:dsprites\_label\_orientation & P & R & 525{,}191 & 58{,}628 & 73{,}728 & 3{,}112/2{,}536/357 & 100.0 & 96.34 & 95.82 & 1000 \\
    kitti\_distance & 50 & cbv:kitti\_closest\_vehicle\_distance & P & P$^\dagger$ & 5{,}712 & 423 & 711 & 0/0/0 & 5.5 & 53.16 & 47.68 & 423 \\
    caltech256 & 50 & ilee0022/Caltech-256 & P & R & 22{,}311 & 2{,}479 & 3{,}061 & 1/0/0 & 0.6 & 96.70 & 91.70 & 1000 \\
    intel\_scenes & 50 & sfarrukhm/intel-image-classification & P & R & 12{,}626 & 1{,}402 & 3{,}000 & 3/3/0 & 0.9 & 96.27 & 95.60 & 1000 \\
    fruits360 & 50 & PedroSampaio/fruits-360 & P & R & 60{,}926 & 6{,}764 & 22{,}688 & 0/0/0 & 52.2 & 99.81 & 99.43 & 1000 \\
    inat\_wildflowers25 & 50 & meganariley/inat-wildflowers-25 & P & R & 7{,}199 & 800 & 1{,}000 & 1/0/0 & 0.0 & 98.00 & 94.00 & 800 \\
    ip102\_insects & 50 & EnmmmmOvO/insect-pest-dataset & P & R & 40{,}589 & 4{,}504 & 22{,}619 & 1/0/1 & 6.3 & 75.04 & 67.07 & 1000 \\
    pokemon & 50 & fcakyon/pokemon-classification & S & R & 5{,}070 & 570 & 1{,}351 & 0/0/0 & 0.1 & 97.93 & 93.56 & 570 \\
    plantvillage & 50 & BrandonFors/Plant-Diseases-PlantVillage-Dataset & P & R & 39{,}100 & 4{,}345 & 10{,}849 & 7/4/0 & 0.2 & 99.82 & 99.59 & 1000 \\
    tomato\_leaf & 50 & wellCh4n/tomato-leaf-disease-image & V & R & 12{,}787 & 1{,}422 & 3{,}569 & 5/4/0 & 0.1 & 99.94 & 99.55 & 1000 \\
    alzheimer\_mri & 50 & Falah/Alzheimer\_MRI & P & R & 4{,}608 & 512 & 1{,}280 & 0/0/0 & 99.3 & 94.06 & 97.34 & 512 \\
    brain\_hemorrhage & 50 & Ahmed-ibn-Harun/BrainHermorrhage & P & R & 6{,}540 & 724 & 811 & 28/0/6 & 16.1 & 88.29 & 82.98 & 724 \\
    dermamnist & 50 & MedMNIST+ (Zenodo 10519652) & P & P & 7{,}007 & 1{,}003 & 2{,}005 & 0/0/0 & 1.6 & 91.87 & 86.13 & 1000 \\
    eeg\_seizure\_spectrogram & 50 & JLB-JLB/seizure\_detection\_224x224\_raw\_frequency & P & P$^\dagger$ & 83{,}765 & 20{,}893 & 31{,}384 & 42/9/4 & 0.1 & 78.60 & 78.87 & 1000 \\
    lc25000 & 50 & Alejandro98/LC25000 & P & R & 15{,}375 & 1{,}730 & 3{,}750 & 252/123/20 & 1.0 & 99.95 & 99.76 & 1000 \\
    hyper\_kvasir & 50 & sahilur/hyper-kvasir-labeled-images & P & P & 8{,}528 & 1{,}069 & 1{,}065 & 0/0/0 & 4.1 & 91.27 & 89.95 & 1000 \\
    organcmnist & 50 & MedMNIST+ (Zenodo 10519652) & P & P & 12{,}975 & 2{,}392 & 8{,}216 & 0/0/0 & 0.2 & 95.61 & 95.93 & 1000 \\
    retinamnist & 50 & MedMNIST+ (Zenodo 10519652) & P & P & 1{,}080 & 120 & 400 & 0/0/0 & 63.1 & 69.00 & 61.75 & 120 \\
    rsscn7 & 50 & jonathan-roberts1/RSSCN7 & H & R & 2{,}016 & 224 & 560 & 0/0/0 & 0.0 & 96.25 & 95.18 & 224 \\
    uc\_merced & 50 & blanchon/UC\_Merced & H & R & 1{,}505 & 166 & 429 & 0/0/0 & 0.3 & 99.30 & 98.83 & 166 \\
    domainnet\_clipart & 50 & wltjr1007/DomainNet & P & R & 30{,}144 & 3{,}354 & 14{,}604 & 21/3/3 & 4.5 & 87.33 & 80.91 & 1000 \\
    domainnet\_sketch & 50 & wltjr1007/DomainNet & P & R & 43{,}393 & 4{,}815 & 20{,}916 & 1/1/2 & 3.6 & 82.69 & 73.01 & 1000 \\
    asl\_hand\_gestures & 50 & juanjodurillo/asl-hg & P & R & 25{,}647 & 2{,}859 & 7{,}200 & 206/67/21 & 91.8 & 99.99 & 99.96 & 1000 \\
    clothes & 50 & samokosik/clothes\_simplifiedv2 & P & R & 3{,}497 & 390 & 686 & 0/0/0 & 0.9 & 96.50 & 94.17 & 390 \\
    football\_players & 50 & aaronqg/golden-foot-football-players & P & R & 4{,}658 & 517 & 719 & 0/0/0 & 8.2 & 96.52 & 94.99 & 517 \\
    human\_action & 50 & Bingsu/Human\_Action\_Recognition & S & R & 8{,}957 & 995 & 2{,}537 & 0/0/0 & 0.1 & 93.81 & 88.96 & 995 \\
    wikiart\_style & 50 & huggan/wikiart & H & R & 58{,}420 & 6{,}500 & 16{,}197 & 219/85/23 & 0.8 & 76.13 & 66.94 & 1000 \\
    yoga\_poses & 50 & AdityasArsenal/Yoga-pose-Data-Set & P & R & 1{,}542 & 176 & 321 & 64/26/5 & 13.1 & 99.38 & 99.69 & 176 \\
    \bottomrule
  \end{tabular}
\end{table*}

\begin{table*}[!tp]
  \ContinuedFloat
  \centering\scriptsize
  \setlength{\tabcolsep}{2.6pt}
  \caption{\textbf{CLIPMerge-100 tasks, provenance and expert accuracies, continued.} Tasks 51--100. Column definitions and footnote symbols follow the first part of Table~\ref{tab:app-provenance}.}
  \label{tab:app-provenance-b}
  \begin{tabular}{@{}lclccrrrrrrrr@{}}
    \toprule
    & & & \multicolumn{2}{c}{Split rule} & \multicolumn{3}{c}{Split size} & & & \multicolumn{2}{c}{Expert acc.} & \\
    \cmidrule(lr){4-5}\cmidrule(lr){6-8}\cmidrule(lr){11-12}
    Task & Set & Source & Test & Val & Train & Val & Test & Dups & ND & L/14 & B/32 & Sub. \\
    \midrule
    fgvc\_aircraft & 100 & cb:fgvc\_aircraft & P & R & 3{,}034 & 300 & 3{,}333 & 0/0/0 & 0.1 & 58.90 & 36.27 & 300 \\
    country211 & 100 & cb:country211 & P & R & 28{,}485 & 3{,}165 & 21{,}100 & 0/0/0 & 1.0 & 36.16 & 18.71 & 1000 \\
    clevr\_count & 100 & cbv:clevr\_count\_all & P & R & 56{,}700 & 6{,}300 & 15{,}000 & 0/0/0 & 0.5 & 96.00 & 91.67 & 1000 \\
    dmlab & 100 & cbv:dmlab & P & P$^\dagger$ & 58{,}994 & 22{,}628 & 22{,}735 & 0/0/0 & 0.1 & 78.04 & 71.02 & 1000 \\
    dsprites\_location & 100 & cbv:dsprites\_label\_x\_position & P & R & 525{,}174 & 58{,}601 & 73{,}728 & 3{,}088/2{,}580/381 & 100.0 & 99.99 & 99.98 & 1000 \\
    smallnorb\_azimuth & 100 & cbv:smallnorb\_label\_azimuth & P & G$^\ddagger$ & 19{,}440 & 4{,}860 & 12{,}150 & 0/0/0 & 37.6 & 27.64 & 26.27 & 1000 \\
    smallnorb\_elevation & 100 & cbv:smallnorb\_label\_elevation & P & G$^\ddagger$ & 19{,}440 & 4{,}860 & 12{,}150 & 0/0/0 & 37.5 & 66.55 & 45.05 & 1000 \\
    diabetic\_retinopathy & 100 & cbv:diabetic\_retinopathy & P & R & 31{,}614 & 3{,}512 & 42{,}670 & 0/0/0 & 0.2 & 80.57 & 75.73 & 1000 \\
    mit\_indoor67 & 100 & keremberke/indoor-scene-classification & S & R & 11{,}111 & 1{,}236 & 3{,}203 & 0/0/0 & 0.3 & 92.57 & 88.95 & 1000 \\
    imagewoof & 100 & leandrodevai/imagewoof-320px-resplit & P & R & 8{,}122 & 901 & 1{,}179 & 1/0/1 & 0.0 & 96.01 & 93.30 & 901 \\
    tiny\_imagenet & 100 & slegroux/tiny-imagenet-200-clean & P & R & 88{,}352 & 9{,}822 & 4{,}923 & 2/3/0 & 0.1 & 89.25 & 79.30 & 1000 \\
    cub200 & 100 & bentrevett/caltech-ucsd-birds-200-2011 & P & R & 5{,}394 & 600 & 5{,}794 & 0/0/0 & 0.1 & 84.97 & 60.86 & 600 \\
    stanford\_dogs & 100 & amaye15/stanford-dogs & P & R & 12{,}911 & 1{,}441 & 6{,}174 & 46/4/4 & 0.1 & 87.58 & 74.81 & 1000 \\
    birds525 & 100 & chriamue/bird-species-dataset & P & P & 84{,}635 & 2{,}625 & 2{,}625 & 0/0/0 & 0.0 & 99.54 & 97.71 & 1000 \\
    butterflies100 & 100 & ManLuML/onmanifold-tfg-butterfly-benchmark & P & R & 3{,}995 & 446 & 170 & 0/0/0 & 0.0 & 97.65 & 97.06 & 446 \\
    whu\_rs19 & 100 & jonathan-roberts1/WHU-RS19 & H & R & 724 & 80 & 201 & 0/0/0 & 0.0 & 100.00 & 99.00 & 80 \\
    vietnamese\_food & 100 & TuyenTrungLe/vietnamese\_food\_images & P & R & 15{,}793 & 1{,}755 & 5{,}040 & 25/6/2 & 0.4 & 93.91 & 86.17 & 1000 \\
    indian\_food & 100 & N16069/indian\_food\_images & P & R & 4{,}774 & 532 & 941 & 16/6/0 & 0.0 & 96.49 & 94.26 & 532 \\
    food11 & 100 & cv-talks/food-11 & P & R & 7{,}418 & 824 & 2{,}750 & 6/1/1 & 0.0 & 99.24 & 97.56 & 824 \\
    sports100 & 100 & HES-XPLAIN/SportsImageClassification & P & P & 13{,}492 & 500 & 500 & 0/0/0 & 0.0 & 100.00 & 99.20 & 500 \\
    beans & 100 & AI-Lab-Makerere/beans & P & R & 931 & 103 & 128 & 0/0/0 & 0.0 & 96.88 & 92.19 & 103 \\
    cassava & 100 & iCassava 2019 zip & P & R & 5{,}076 & 564 & 1{,}885 & 12/2/2 & 0.1 & 90.72 & 85.73 & 564 \\
    cotton\_leaf & 100 & YaswanthReddy23/Cotton\_leaf & P & R & 1{,}342 & 150 & 218 & 0/1/0 & 0.0 & 96.79 & 97.71 & 150 \\
    fruit\_ripeness & 100 & darthraider/fruit-ripeness-detection-dataset & P & R & 3{,}599 & 400 & 1{,}000 & 0/0/0 & 0.1 & 100.00 & 100.00 & 400 \\
    colorectal\_histology & 100 & dpdl/colorectal\_histology & H & R & 3{,}616 & 401 & 983 & 0/0/0 & 0.0 & 96.34 & 95.52 & 401 \\
    malaria\_cells & 100 & dpdl/malaria & H & R & 19{,}856 & 2{,}206 & 5{,}496 & 0/0/0 & 12.6 & 97.73 & 97.09 & 1000 \\
    breastmnist & 100 & MedMNIST+ (Zenodo 10519652) & P & P & 545 & 78 & 156 & 1/0/0 & 8.4 & 91.03 & 91.67 & 78 \\
    bloodmnist & 100 & MedMNIST+ (Zenodo 10519652) & P & P & 11{,}951 & 1{,}711 & 3{,}421 & 5/3/1 & 0.0 & 99.15 & 98.80 & 1000 \\
    tissuemnist & 100 & MedMNIST+ (Zenodo 10519652) & P & P & 165{,}466 & 23{,}640 & 47{,}280 & 0/0/0 & 0.7 & 77.17 & 74.22 & 1000 \\
    organamnist & 100 & MedMNIST+ (Zenodo 10519652) & P & P & 34{,}561 & 6{,}491 & 17{,}778 & 0/0/0 & 0.5 & 97.89 & 97.19 & 1000 \\
    organsmnist & 100 & MedMNIST+ (Zenodo 10519652) & P & P & 13{,}932 & 2{,}452 & 8{,}827 & 0/0/0 & 0.1 & 85.42 & 84.40 & 1000 \\
    brain\_tumor\_mri & 100 & Simezu/brain-tumour-MRI-scan & P & R & 4{,}940 & 550 & 1{,}311 & 170/30/22 & 27.1 & 99.08 & 96.72 & 550 \\
    chest\_xray\_covid & 100 & Ajay-user/pneumonia-detection-chest-xray & P & R & 4{,}622 & 515 & 1{,}288 & 0/7/0 & 57.0 & 97.05 & 96.74 & 515 \\
    dermnet & 100 & exper1ment/dermnet & P & R & 13{,}353 & 1{,}499 & 4{,}002 & 520/128/57 & 9.5 & 71.06 & 55.25 & 1000 \\
    patternnet & 100 & blanchon/PatternNet & H & R & 21{,}828 & 2{,}424 & 6{,}147 & 1/0/0 & 1.0 & 99.79 & 99.66 & 1000 \\
    domainnet\_infograph & 100 & wltjr1007/DomainNet & P & R & 32{,}422 & 3{,}601 & 15{,}582 & 0/0/0 & 2.9 & 67.23 & 53.26 & 1000 \\
    domainnet\_painting & 100 & wltjr1007/DomainNet & P & R & 45{,}374 & 5{,}042 & 21{,}850 & 0/0/0 & 1.3 & 84.47 & 75.68 & 1000 \\
    domainnet\_quickdraw & 100 & wltjr1007/DomainNet & P & R & 108{,}675 & 12{,}075 & 51{,}750 & 0/0/0 & 1.9 & 72.69 & 67.60 & 1000 \\
    domainnet\_real & 100 & wltjr1007/DomainNet & P & R & 108{,}821 & 12{,}082 & 52{,}041 & 1/1/1 & 4.6 & 91.96 & 87.37 & 1000 \\
    usps & 100 & flwrlabs/usps & P & R & 6{,}564 & 727 & 2{,}007 & 0/0/0 & 35.8 & 98.26 & 97.91 & 727 \\
    bangla\_vs\_assamese\_script & 100 & anrikus/lexical\_diff\_bangla\_assamese\_v2 & P & R & 609 & 70 & 300 & 19/2/0 & 80.8 & 81.00 & 75.00 & 70 \\
    rvl\_cdip & 100 & chainyo/rvl-cdip & P & R & 287{,}266 & 31{,}937 & 39{,}999 & 594/139/63 & 9.2 & 92.70 & 89.57 & 1000 \\
    galaxy10\_decals & 100 & matthieulel/galaxy10\_decals & P & R & 14{,}346 & 1{,}598 & 1{,}774 & 7/10/1 & 1.0 & 85.91 & 83.60 & 1000 \\
    fairface\_age & 100 & HuggingFaceM4/FairFace & V & R & 78{,}023 & 8{,}672 & 10{,}954 & 19/28/2 & 1.3 & 62.55 & 59.01 & 1000 \\
    wonders\_of\_world & 100 & Saul08/wonders-of-world-classification & P & R & 2{,}422 & 270 & 381 & 0/0/0 & 6.9 & 99.74 & 99.48 & 270 \\
    hagrid\_gestures & 100 & ntsrigaud/hagrid-subset & P & P & 14{,}591 & 1{,}728 & 2{,}880 & 1/0/0 & 0.0 & 97.74 & 87.67 & 1000 \\
    deepfake\_vs\_real & 100 & itsLeen/deepfake\_vs\_real\_image\_detection & P & R & 5{,}694 & 633 & 1{,}117 & 0/0/0 & 0.0 & 99.46 & 98.48 & 633 \\
    music\_genre\_spectrogram & 100 & ghermoso/egtzan\_plus & P & R & 1{,}522 & 168 & 189 & 3/4/0 & 0.2 & 70.90 & 76.72 & 168 \\
    chess\_pieces & 100 & S1M0N38/chess-cv-chessvision & V & R & 2{,}334 & 257 & 595 & 0/0/0 & 1.0 & 100.00 & 100.00 & 257 \\
    person\_presence & 100 & Harvard-Edge/Wake-Vision & P & P$^\dagger$ & 180{,}000 & 18{,}581 & 55{,}762 & 0/0/0 & 0.3 & 87.79 & 87.35 & 1000 \\
    \bottomrule
  \end{tabular}
\end{table*}

%% file: sec/app_native_grids.tex
\begin{table*}[t]
  \centering\footnotesize
  \setlength{\tabcolsep}{4pt}
  \caption{\textbf{Exact Native search grids.} Both backbones use these grids at every task count. Products enumerate all combinations, whereas unions add only the stated points. Here, $s$ denotes the method's output scale, $n$ the number of samples used to estimate statistics, and $r$ the RegMean off-diagonal shrinkage factor, denoted by $\alpha$ in Sec.~\ref{sec:rw}. The column $c$ gives the number of candidate configurations.}
  \label{tab:app-native-grids}
  \parbox{\textwidth}{
  $\mathcal S=\{0.01,0.02,0.03,0.05,0.07,0.1,0.15,0.2,0.3,0.4,0.5,0.7,1\}$;\quad
  $\mathcal R=\{0.5,0.6,0.7,0.8,0.9,0.95,1\}$.}
  \par\smallskip
  \begin{tabular}{@{}l>{\raggedright\arraybackslash}p{0.75\textwidth}r@{}}
    \toprule
    Method & Candidate configurations & $c$\\
    \midrule
    Task Arithmetic & $s\in\mathcal S$. & 13\\
    Fisher & $(n,\text{normalize})\in\{64,128,256,512,1024\}\times\{\text{true},\text{false}\}$. & 10\\
    TIES & Sum: $(k,s)\in\{10,20,30\}\times\mathcal S\;\cup\;\{5\}\times\{0.1,0.15,0.2,0.3\}$. Mean: $(k,s)\in\{10,20,30\}\times\{0.5,0.7,1,1.3,1.6,2,2.5,3\}\;\cup\;\{5\}\times\{0.5,0.7,1,1.3\}$. Here $k$ is the trimming threshold. & 71\\
    TSV-M & $\alpha\in\{0.1,0.2,0.3,0.5,0.7,0.8,0.9,1,1.1,1.2,1.4,1.6,2,2.5,3\}$. & 15\\
    ISO-CTS & $(\rho,s)\in\{0.1,0.2,\dots,0.9\}\times\{0.5,1,1.5,2,2.5,3\}\;\cup\;\{0.95\}\times\{0.5,1,1.5\}$, where $\rho$ is the common-space fraction. & 57\\
    RegMean & $(n,r)\in\{128,256,512,1024\}\times\mathcal R$. & 28\\
    RegMean++ & $(n,r)\in\{256,512\}\times\mathcal R\;\cup\;\{128,1024\}\times\{0.9,0.95\}$. & 18\\
    LOT & $s\in\{0.5,0.7,0.9,1,1.1,1.2,1.3,1.5,1.7,2\}$; 64 exemplars, exemplar seed 0. & 10\\
    WUDI & Iterations $\in\{100,200,400,800,1600,3200,6400,12800\}$. & 8\\
    ACE-Merging & $(\epsilon,k,s)\in\{10^{-7},10^{-6},10^{-5},10^{-4},10^{-3},10^{-2}\}\times\{0.1,0.2,0.3,0.4,0.5,0.6\}\times\{0.5,0.75,1,1.25,1.5,2\}$; $k$ is the retained-rank fraction. & 216\\
    \bottomrule
  \end{tabular}
\end{table*}

%% file: sec/tab_head80.tex
\begin{table*}[!tp]
  \centering
  \caption{\textbf{Retention on fixed head and tail task sets.} Tasks are partitioned using Native's test retention, and the partition is held fixed when evaluating +\RaTW. head80 is the mean retention (\%) over the 80\% of tasks outside Native's tail20 set. $\Delta$ denotes +\RaTW minus Native in percentage points, computed before rounding. $\Delta$tail20 reports the change in mean retention on the complementary fixed 20\% set; it can differ from the tail20 gains in Tables~\ref{tab:transfer} and~\ref{tab:transfer-b32}, where the tail20 set is recomputed for each model. Average is taken over all eleven methods within each backbone and task count.}
  \label{tab:app-head80}
  \footnotesize
  \renewcommand{\arraystretch}{1.08}
  \setlength{\aboverulesep}{0.3ex}\setlength{\belowrulesep}{0.5ex}
  \setlength{\tabcolsep}{1pt}
  \begin{tabular*}{\textwidth}{@{\extracolsep{\fill}}cl*{12}{r}@{}}
    \toprule
    & & \multicolumn{4}{c}{$T=20$} & \multicolumn{4}{c}{$T=50$} & \multicolumn{4}{c}{$T=100$} \\
    \cmidrule(lr){3-6}\cmidrule(lr){7-10}\cmidrule(l){11-14}
    & & \multicolumn{3}{c}{head80} & \multicolumn{1}{c}{tail20} & \multicolumn{3}{c}{head80} & \multicolumn{1}{c}{tail20} & \multicolumn{3}{c}{head80} & \multicolumn{1}{c}{tail20} \\
    \cmidrule(lr){3-5}\cmidrule(lr){6-6}\cmidrule(lr){7-9}\cmidrule(lr){10-10}\cmidrule(lr){11-13}\cmidrule(l){14-14}
    & Method & \multicolumn{1}{c}{Native} & \multicolumn{1}{c}{+\RaTW} & \multicolumn{1}{c}{$\Delta$} & \multicolumn{1}{c}{$\Delta$} & \multicolumn{1}{c}{Native} & \multicolumn{1}{c}{+\RaTW} & \multicolumn{1}{c}{$\Delta$} & \multicolumn{1}{c}{$\Delta$} & \multicolumn{1}{c}{Native} & \multicolumn{1}{c}{+\RaTW} & \multicolumn{1}{c}{$\Delta$} & \multicolumn{1}{c}{$\Delta$} \\
    \midrule
    \multicolumn{14}{@{}l}{\textit{ViT-L/14}}\\
    \midrule
    \multirow{5}{*}{\rotatebox[origin=c]{90}{\scriptsize Aggr.}} & Task Arithmetic & 86.8 & 84.4 & $-2.4$ & $+23.4$ & 75.1 & 74.4 & $-0.7$ & $+6.9$ & 67.4 & 67.0 & $-0.4$ & $+2.7$ \\
     & Fisher & 85.5 & 81.6 & $-3.9$ & $+20.5$ & 73.3 & 72.5 & $-0.7$ & $+7.6$ & 66.1 & 66.0 & $-0.2$ & $+7.9$ \\
     & TIES & 87.9 & 86.0 & $-1.9$ & $+26.5$ & 75.3 & 75.3 & $+0.0$ & $+2.7$ & 68.0 & 67.9 & $-0.1$ & $+0.4$ \\
     & TSV-M & 95.4 & 95.3 & $-0.0$ & $+11.0$ & 88.4 & 86.2 & $-2.2$ & $+21.4$ & 82.0 & 81.8 & $-0.2$ & $+5.2$ \\
     & ISO-CTS & 96.4 & 97.0 & $+0.6$ & $+5.3$ & 91.6 & 90.9 & $-0.7$ & $+20.6$ & 85.3 & 86.6 & $+1.3$ & $+6.8$ \\
    \cmidrule(l){2-14}
    \multirow{6}{*}{\rotatebox[origin=c]{90}{\scriptsize Recon.}} & RegMean & 94.3 & 93.3 & $-1.0$ & $+19.7$ & 85.6 & 84.3 & $-1.3$ & $+21.8$ & 77.7 & 76.5 & $-1.2$ & $+9.8$ \\
     & RegMean++ & 94.5 & 94.3 & $-0.1$ & $+15.7$ & 86.6 & 85.5 & $-1.0$ & $+20.7$ & 80.0 & 78.9 & $-1.1$ & $+9.2$ \\
     & LOT & 94.1 & 92.7 & $-1.4$ & $+20.5$ & 86.3 & 84.1 & $-2.2$ & $+22.4$ & 78.8 & 78.7 & $-0.1$ & $+5.4$ \\
     & WUDI & 97.0 & 96.8 & $-0.2$ & $+9.6$ & 90.8 & 89.5 & $-1.3$ & $+25.6$ & 83.8 & 83.6 & $-0.2$ & $+7.4$ \\
     & ACE-Merging & 96.7 & 96.1 & $-0.7$ & $+13.2$ & 88.5 & 88.2 & $-0.3$ & $+21.9$ & 79.5 & 77.4 & $-2.1$ & $+17.8$ \\
     & ReForge & 98.0 & 97.8 & $-0.2$ & $+5.0$ & 95.7 & 95.1 & $-0.6$ & $+8.3$ & 90.7 & 90.4 & $-0.3$ & $+8.6$ \\
    \cmidrule(l){2-14}
    \multicolumn{2}{@{}l}{\textit{Average}} & 93.3 & 92.3 & $-1.0$ & $+15.5$ & 85.2 & 84.2 & $-1.0$ & $+16.3$ & 78.1 & 77.7 & $-0.4$ & $+7.4$ \\
    \midrule
    \multicolumn{14}{@{}l}{\textit{ViT-B/32}}\\
    \midrule
    \multirow{5}{*}{\rotatebox[origin=c]{90}{\scriptsize Aggr.}} & Task Arithmetic & 79.0 & 75.7 & $-3.4$ & $+10.1$ & 66.9 & 66.1 & $-0.7$ & $+3.0$ & 60.6 & 60.4 & $-0.2$ & $+0.7$ \\
     & Fisher & 80.3 & 74.7 & $-5.6$ & $+24.0$ & 67.8 & 66.8 & $-1.0$ & $+4.5$ & 60.8 & 61.1 & $+0.3$ & $+2.2$ \\
     & TIES & 80.6 & 75.9 & $-4.7$ & $+17.3$ & 67.5 & 66.0 & $-1.5$ & $+7.0$ & 61.6 & 61.1 & $-0.5$ & $+2.0$ \\
     & TSV-M & 89.3 & 86.9 & $-2.4$ & $+19.6$ & 81.5 & 80.6 & $-0.9$ & $+9.3$ & 72.9 & 72.6 & $-0.3$ & $+2.4$ \\
     & ISO-CTS & 89.5 & 86.3 & $-3.3$ & $+18.3$ & 83.8 & 83.7 & $-0.0$ & $+8.4$ & 75.9 & 75.9 & $+0.0$ & $+3.5$ \\
    \cmidrule(l){2-14}
    \multirow{6}{*}{\rotatebox[origin=c]{90}{\scriptsize Recon.}} & RegMean & 88.6 & 86.9 & $-1.8$ & $+16.6$ & 78.0 & 77.1 & $-0.9$ & $+11.0$ & 69.6 & 69.1 & $-0.4$ & $+4.5$ \\
     & RegMean++ & 90.5 & 88.4 & $-2.1$ & $+22.5$ & 81.7 & 80.1 & $-1.5$ & $+18.8$ & 74.0 & 73.6 & $-0.4$ & $+5.9$ \\
     & LOT & 87.0 & 86.2 & $-0.8$ & $+11.4$ & 77.7 & 75.8 & $-1.9$ & $+11.3$ & 68.9 & 68.3 & $-0.6$ & $+4.5$ \\
     & WUDI & 90.5 & 90.1 & $-0.4$ & $+12.6$ & 82.0 & 81.4 & $-0.7$ & $+9.0$ & 73.8 & 73.2 & $-0.6$ & $+3.4$ \\
     & ACE-Merging & 89.9 & 90.2 & $+0.2$ & $+15.1$ & 78.5 & 75.5 & $-3.0$ & $+27.5$ & 70.0 & 69.9 & $-0.1$ & $+7.0$ \\
     & ReForge & 93.8 & 93.8 & $-0.0$ & $+11.1$ & 88.5 & 87.6 & $-1.0$ & $+9.4$ & 81.5 & 80.8 & $-0.7$ & $+4.0$ \\
    \cmidrule(l){2-14}
    \multicolumn{2}{@{}l}{\textit{Average}} & 87.2 & 85.0 & $-2.2$ & $+16.2$ & 77.6 & 76.4 & $-1.2$ & $+10.8$ & 70.0 & 69.6 & $-0.3$ & $+3.7$ \\
    \bottomrule
  \end{tabular*}
\end{table*}

%% file: sec/app_vlm.tex
\section{Vision-Language Protocol and Additional Results}
\label{app:vlm}
This section describes the VLM protocol used in Sec.~\ref{sec:vlm} and reports the full results in Table~\ref{tab:app-vlm-summary}.

\input{sec/tab_vlm_main}

\paragraph{Setup.}
\label{app:vlm-setup}\label{app:vlm-protocol}%
The ten tasks comprise ScienceQA-IMG, IconQA, VSR, TallyQA, PathVQA, and VizWiz, together with four classification tasks: Stanford Cars, FGVC-Aircraft, DTD, and RESISC45. Each classification task is presented as four-way multiple choice. InternVL2.5-1B and Qwen2-VL-7B use rank-16 LoRA experts on language-model projections, with the vision encoder and projector frozen.

Calibration, validation, and test blocks are separated by image group. Unlike the CLIP experiments, both VLMs estimate recovery scores on calibration data, combining two 256-question blocks into a set of 512 questions per task, denoted by $\mathcal C_t$. \DAs activation statistics use 128 questions per task (1,280 total), with a maximum sequence length of 8,192. Both models use 17,871 validation questions for model selection and 32,421 test questions for evaluation. Answers are decoded greedily and evaluated with each task's fixed scoring rule; at ten tasks, tail20 averages retention over the two worst-retained tasks.

Each answer receives a task-specific score $s(x)\in[0,1]$, allowing a continuous recovery count:
\[
k_t=\sum_{x\in\mathcal C_t}\max\bigl(0,s_{E_t}(x)-s_R(x)\bigr).
\]
This reduces to the CLIP count in Eq.~\eqref{eq:beta} when scores are binary. We use the same floor, $\beta_t=\max(k_t,1)/|\mathcal C_t|$, and the weight mapping in Eq.~\eqref{eq:q}, although the floor now also applies when $0<k_t<1$. Since every task has $|\mathcal C_t|=512$, the recovery-score floor is identical across tasks, and the weights are uniform whenever all $k_t\leq1$.

For Native Task Arithmetic, TIES, and OptMerge, we search output scales $\{0.1,0.3,0.5,0.7,1,1.5\}$; +\RaTW searches their Cartesian product with $\gamma\in\{0,0.25,0.5,0.75,1,1.5,2,3\}$. Each +\RaTW variant therefore evaluates 48 candidates, comprising six reused Native candidates at $\gamma=0$ and 42 additional candidates.

On InternVL2.5-1B, ReForge and \RFpp use Native TIES as the initial reference and anchor for \DFs, and Native OptMerge for \DAs. On Qwen2-VL-7B, both variants use Native TIES.

Each refinement round uses 50 completed BO trials; failed or pruned proposals do not count toward this budget. The search covers an output scale in $[1,1.3]$ and four log-scaled ridge coefficients in $[10^{-3},100]$, corresponding to attention input, attention output, MLP input, and MLP output projections; weighted variants also search the eight $\gamma$ values above. BO uses GPSampler with seed 0 and ten startup trials, with one seeded random proposal after 25 consecutive duplicate or pruned proposals. Model selection maximizes the equally weighted ten-task validation mean, breaking BO ties in favor of the smaller trial number. Merged task vectors are materialized in float32 before a single bfloat16 cast. In subsequent \RFpp rounds, the preceding round's selected model becomes both the anchor and the reference for recomputing recovery scores. The three rounds use 150 completed BO trials in total. This budget excludes initial-model preparation, expert/reference scoring, activation collection, and final testing.

\paragraph{Scope.}
\label{app:vlm-disclosures}
All VLM BO searches use seed 0, and the experts of each backbone use one training seed. The validation-set size was determined on InternVL2.5-1B using a selection-stability criterion across two random halves: the candidate selected on either half should lose at most 0.3\,pp of validation mean on the other half relative to that half's own best candidate. For OptMerge, this criterion remained unmet after three planned expansions from 10,000 to 17,871 questions, so selection uses the full expanded set. Qwen2-VL-7B uses the same validation set and meets the criterion.

On InternVL2.5-1B, the ScienceQA and TallyQA experts improve validation performance over the base model by 1.2 and 1.1\,pp, respectively, and five tasks---ScienceQA, IconQA, VSR, TallyQA, and PathVQA---appear in the base model's published training data. The expert learning rate was not re-tuned after the classification tasks took their final multiple-choice format. Prompting and image processing differ between backbones, so their comparison does not isolate model size.

\paragraph{Results.}
Adding \RaTW improves both Mean and tail20 for all five merging methods on both backbones (Table~\ref{tab:app-vlm-summary}). Six of the ten Mean gains are below 0.5\,pp: TIES and ReForge with \DFs on both backbones, and OptMerge and ReForge with \DAs on Qwen2-VL-7B. These results cover OptMerge, which is designed for merging multimodal large language models, as well as both ReForge variants.

Three-round \RFpp with \DAs improves both metrics over its single-round weighted counterpart and achieves the highest values among the tabulated merging configurations on both backbones, using three times the BO trial budget of one-round ReForge. On Qwen2-VL-7B, its Mean is 0.20\,pp below the expert average, and its tail20 reaches 99.1\%. Additional rounds do not always improve performance: on InternVL2.5-1B, three-round \RFpp with \DFs yields 71.39 / 72.9 in Mean / tail20, compared with 71.82 / 73.4 for its single-round weighted counterpart.

%% file: sec/tab_vlm_main.tex
\begin{table*}[!t]
\centering\footnotesize
\newcommand{\vlmpair}[2]{\makebox[23pt][r]{#1}\,/\,\makebox[21pt][r]{#2}}
\caption{\textbf{VLM merging results.} The evaluation covers ten tasks on InternVL2.5-1B and Qwen2-VL-7B, with one seed for BO searches. Entries report Mean / tail20 (\%), and $\Delta$ denotes +\RaTW minus Native in percentage points, computed from unrounded scores. At $T=10$, tail20 averages the retention of the two worst-retained tasks. The final two rows report three-round \RFpp with 150 BO trials, whereas one-round ReForge uses 50. Base model denotes the pre-trained model before LoRA fine-tuning, evaluated with the same prompts and scoring rules. On the Qwen2-VL-7B base model, only 0\%, 1\% and 6\% of answers follow the required format on TallyQA, FGVC-Aircraft and DTD, respectively, so its scores reflect format compliance as well as task ability.}
\label{tab:app-vlm-summary}
\label{tab:app-vlm-q7b-single}
\setlength{\tabcolsep}{3pt}
\renewcommand{\arraystretch}{1.15}
\begin{tabular}{@{}lccc@{\hspace{15pt}}ccc@{}}
\toprule
& \multicolumn{3}{c}{InternVL2.5-1B} & \multicolumn{3}{c}{Qwen2-VL-7B}\\
\cmidrule(lr){2-4}\cmidrule(l){5-7}
Method & Native & +\RaTW & $\Delta$ & Native & +\RaTW & $\Delta$\\
\midrule[0.35pt]
\addlinespace[2pt]
Base model & \vlmpair{59.59}{50.6} & -- & -- & \vlmpair{24.14}{0.0} & -- & --\\
Expert average & \vlmpair{81.77}{100.0} & -- & -- & \vlmpair{87.87}{100.0} & -- & --\\
\midrule[0.35pt]
\addlinespace[2pt]
Task Arithmetic & \vlmpair{68.93}{66.2} & \textbf{\vlmpair{70.62}{74.7}} & \vlmpair{+1.69}{+8.5} & \vlmpair{82.56}{72.2} & \textbf{\vlmpair{83.63}{85.7}} & \vlmpair{+1.07}{+13.5}\\
TIES & \vlmpair{70.31}{70.7} & \textbf{\vlmpair{70.48}{71.7}} & \vlmpair{+0.17}{+1.0} & \vlmpair{84.94}{88.4} & \textbf{\vlmpair{85.18}{89.8}} & \vlmpair{+0.24}{+1.4}\\
OptMerge & \vlmpair{67.74}{63.1} & \textbf{\vlmpair{68.53}{68.8}} & \vlmpair{+0.79}{+5.7} & \vlmpair{85.16}{90.9} & \textbf{\vlmpair{85.66}{92.9}} & \vlmpair{+0.49}{+2.0}\\
ReForge-DF & \vlmpair{71.50}{72.6} & \textbf{\vlmpair{71.82}{73.4}} & \vlmpair{+0.31}{+0.8} & \vlmpair{85.50}{90.6} & \textbf{\vlmpair{85.99}{93.3}} & \vlmpair{+0.48}{+2.7}\\
ReForge-DA & \vlmpair{74.05}{78.3} & \textbf{\vlmpair{74.95}{83.8}} & \vlmpair{+0.90}{+5.4} & \vlmpair{87.10}{96.2} & \textbf{\vlmpair{87.30}{97.7}} & \vlmpair{+0.20}{+1.5}\\
\midrule[0.35pt]
\addlinespace[2pt]
ReForge++-DF & -- & \vlmpair{71.39}{72.9} & -- & -- & \vlmpair{86.99}{93.6} & --\\
\textbf{ReForge++-DA} & -- & \textbf{\vlmpair{77.31}{91.8}} & -- & -- & \textbf{\vlmpair{87.67}{99.1}} & --\\
\bottomrule
\end{tabular}
\end{table*}